\documentclass[sigconf]{acmart}

\usepackage{xspace}
\usepackage{algorithm}
\usepackage{algorithmic}
\usepackage{amsmath}
\usepackage{amsfonts}
\usepackage{amsthm}
\usepackage{subcaption}
\usepackage{booktabs}
\usepackage{multirow}
\usepackage{tabularx}
\usepackage{times}
\usepackage{helvet}
\usepackage{courier}
\usepackage{xcolor}
\usepackage{color}
\usepackage{colortbl}
\usepackage{enumitem}
\usepackage{bm}
\usepackage{natbib}
\usepackage{caption}
\usepackage{courier}
\usepackage{newfloat}
\usepackage{listings}
\usepackage{graphicx}

\definecolor{mygray}{gray}{.9}
\newcommand{\model}{ADAligner\xspace}

\AtBeginDocument{%
  }

\setcopyright{acmlicensed}
\copyrightyear{2018}
\acmYear{2018}
\acmDOI{XXXXXXX.XXXXXXX}
\acmConference[Conference acronym 'XX]{Make sure to enter the correct
  conference title from your rights confirmation email}{June 03--05,
  2018}{Woodstock, NY}
\acmISBN{978-1-4503-XXXX-X/2018/06}

\begin{document}


\title{Learning Noise-Resilient and Transferable Graph-Text Alignment via Dynamic Quality Assessment}

\author{Yuhang Liu}
\affiliation{%
  \institution{School of New Media and Communication, Tianjin University}
  \city{Tianjin}
  \country{China}
}
\email{liuyuhang_13@tju.edu.cn}

\author{Minglai	Shao}
\authornote{Corresponding author.}
\affiliation{%
  \institution{School of New Media and Communication, Tianjin University}
  \city{Tianjin}
  \country{China}}
\email{shaoml@tju.edu.cn}

\author{Zengyi Wo}
\affiliation{%
  \institution{Baidu}
  \city{Beijing}
  \country{China}
}
\email{wozengyi1999@tju.edu.cn}

\author{Yunlong	Chu}
\affiliation{%
  \institution{School of New Media and Communication, Tianjin University}
  \city{Tianjin}
  \country{China}}
\email{2024245030@tju.edu.cn}

\author{Bing Hao}
\affiliation{%
  \institution{School of New Media and Communication, Tianjin University}
  \city{Tianjin}
  \country{China}}
\email{haobing@tju.edu.cn}

\author{Shengzhong	Liu}
\affiliation{%
  \institution{Shanghai Jiao Tong University}
  \city{Shanghai}
  \country{China}}
\email{shengzhong@sjtu.edu.cn}

\author{Ruijie Wang}
\authornotemark[1]
\affiliation{%
  \institution{School of Computer Science and Engineering, Beihang University}
  \city{Beijing}
  \country{China}}
\email{ruijiew@buaa.edu.cn}

\author{Jianxin	Li}
\affiliation{%
  \institution{School of Computer Science and Engineering, Beihang University}
  \city{Beijing}
  \country{China}}
\email{lijx@buaa.edu.cn}

\renewcommand{\shortauthors}{Trovato et al.}


\begin{abstract}
Pre-training Graph Foundation Models (GFMs) on text-attributed graphs (TAGs) is central to web-scale applications such as search, recommendation, and knowledge discovery. However, existing CLIP-style graph–text aligners face two key limitations: they assume strict one-to-one correspondences between nodes and texts, overlooking the inherent many-to-many relations in real-world graphs; and they rely on static alignment objectives that cannot adapt to varying data quality, making them brittle under noisy supervision. Together, these limitations expose a core dilemma: embracing expressive many-to-many alignment amplifies noise, while reverting to strict one-to-one strategies sacrifices semantic diversity and fails to handle inherently mismatched pairs.
To address these challenges, we propose \textbf{\model}, a \emph{dynamic, quality-aware} graph–text alignment framework that dynamically adjusts between expressive many-to-many and conservative one-to-one objectives according to supervision quality. \model estimates batch-level alignment reliability in real time and adapts its optimization accordingly—promoting soft, subgraph-level many-to-many alignment when supervision is clean, while emphasizing reliable one-to-one alignment by dynamically filtering low-confidence pairs under noise. Theoretically, we prove that this dynamic mechanism forms a stable negative feedback process, ensuring convergence and robustness.
Comprehensive experiments on nine diverse TAG datasets demonstrate that \model consistently outperforms prior graph–text aligners on zero-/few-shot node classification, link prediction, and cross-modal retrieval tasks. It maintains strong robustness under noisy supervision and accelerates pre-training by approximately $2$–$3\times$ compared to multimodal baselines, establishing a scalable and reliable foundation for graph–text representation learning in real-world web environments. 
\end{abstract}



\received{20 February 2007}
\received[revised]{12 March 2009}
\received[accepted]{5 June 2009}

\maketitle

\section{Introduction}

\begin{figure}[t]
    \centering 
    \captionsetup{skip=3pt}

    \begin{subfigure}[b]{\columnwidth}
        \centering
        \includegraphics[width=0.9\linewidth]{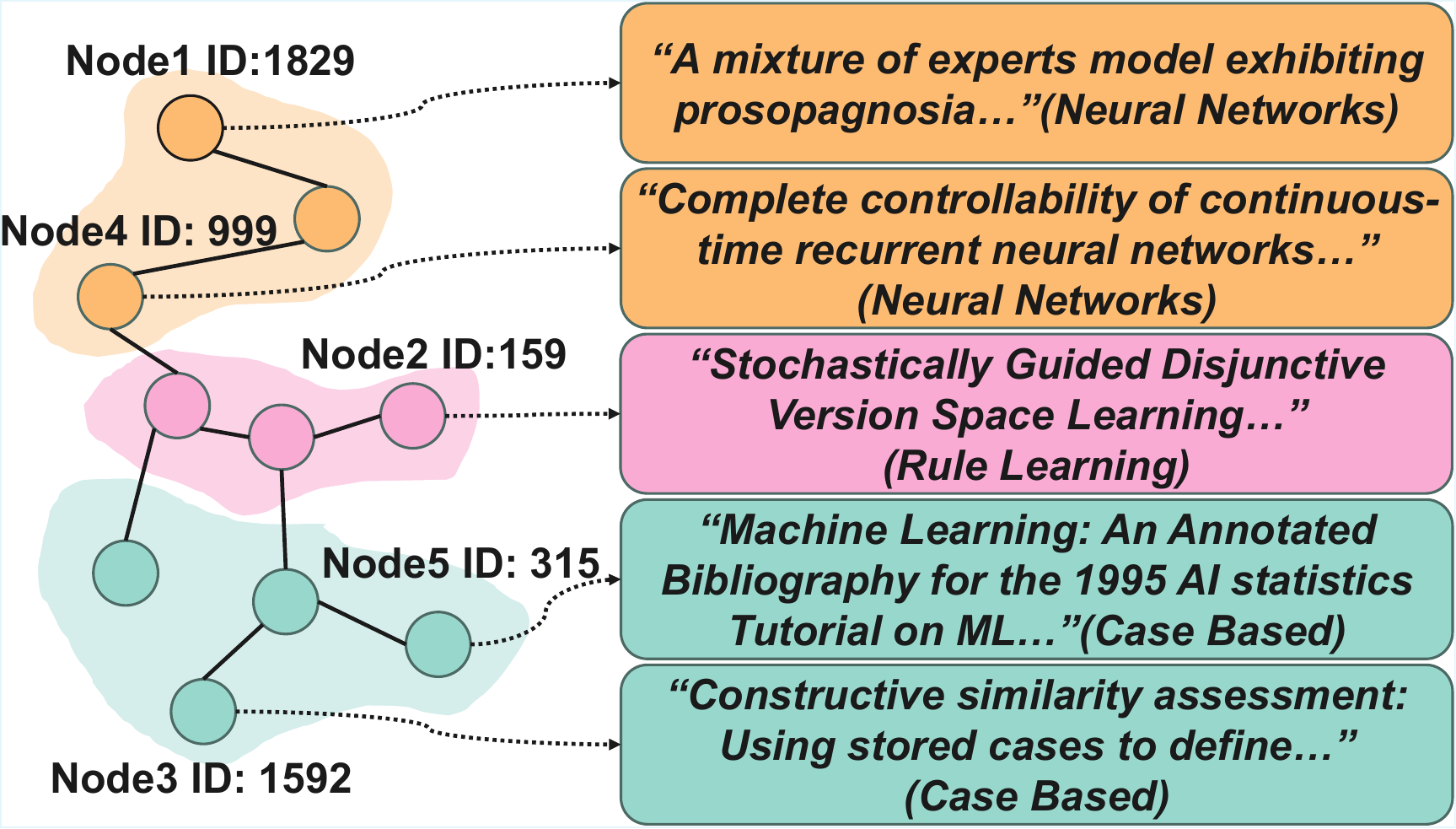}
        \caption{Randomly sampled node-text pairs.}
        \label{fig:sub_text_examples}
    \end{subfigure}

    \begin{subfigure}[b]{0.46\columnwidth}
        \centering
        \includegraphics[width=\linewidth]{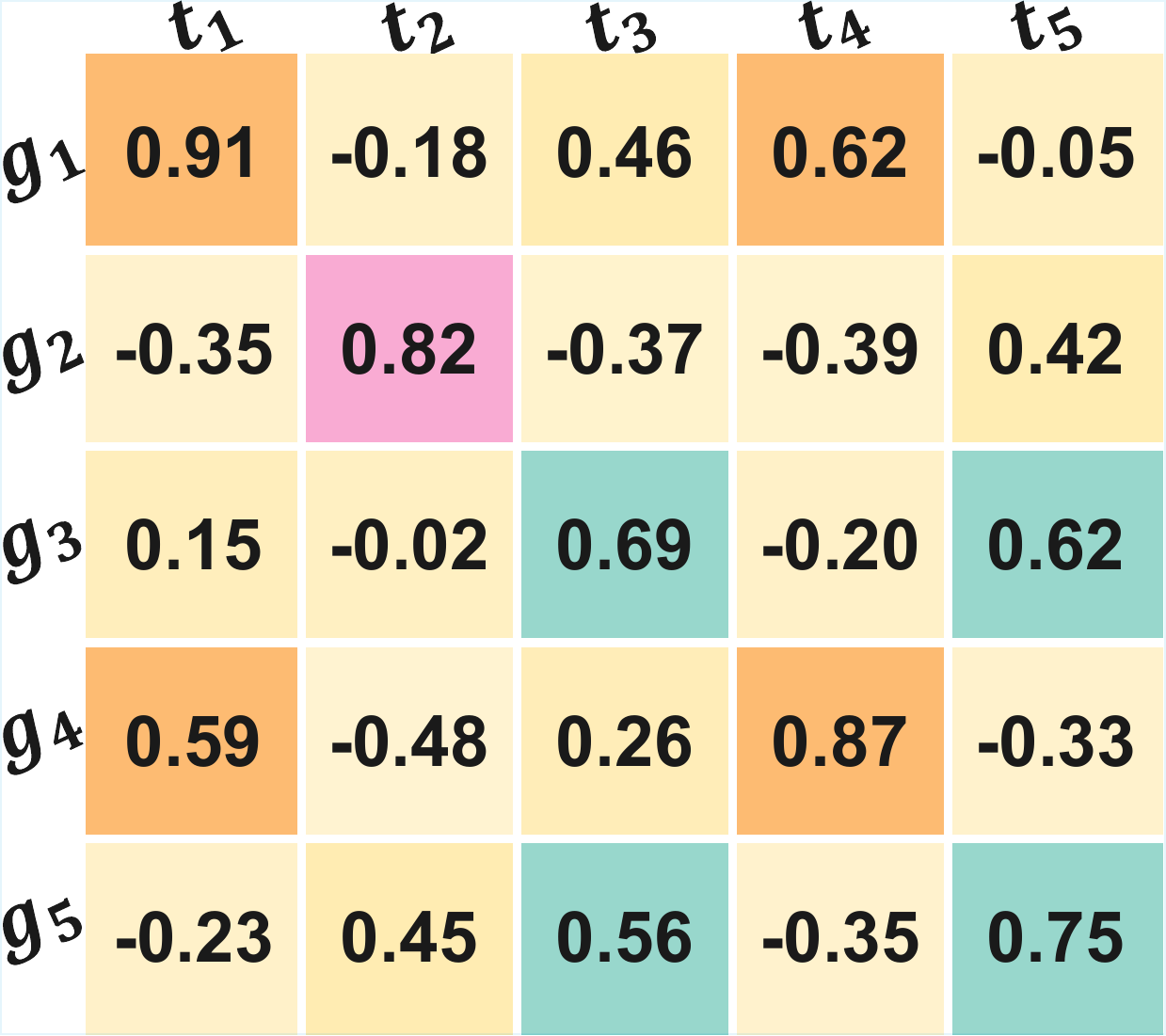}
        \caption{Ideal many-to-many alignment.}
        \label{fig:sub_similarity_matrix}
    \end{subfigure}
    \hfill 
    \begin{subfigure}[b]{0.46\columnwidth}
        \centering
        \includegraphics[width=\linewidth]{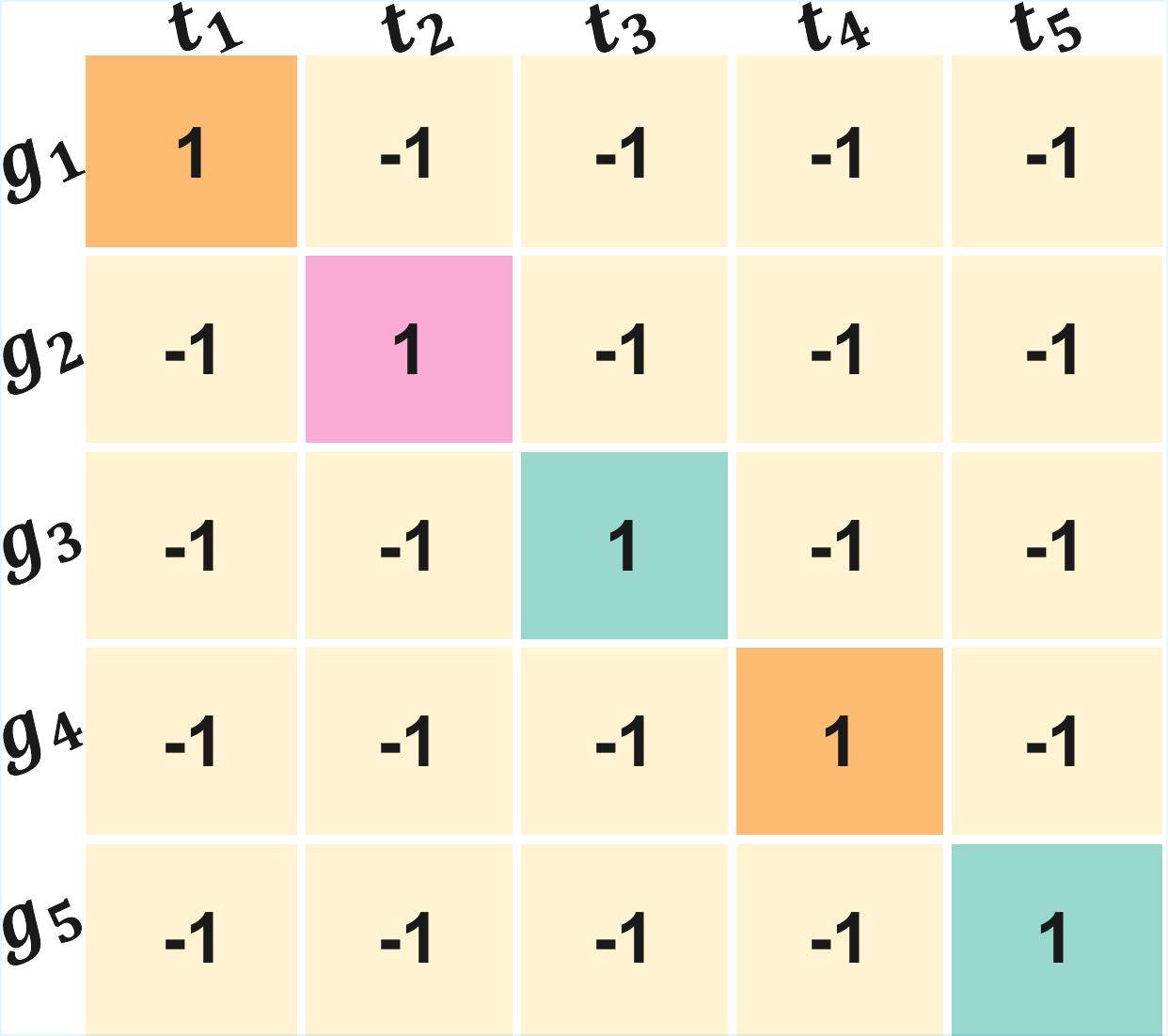}
        \caption{Original one-to-one CLIP alignment.}
        \label{fig:sub_node_categories}
    \end{subfigure}

    \caption{An illustration of many-to-many correspondences on a \textit{Cora} dataset sample. The computed similarity matrix (b) shows complex intra- and inter-class associations, reflecting limitations of original one-to-one alignment (c).}
    \label{fig:motivation_illustrations}
\end{figure}

Graph foundation models (GFMs) have emerged as a unifying paradigm for inductive representation learning in web-scale applications such as recommendation~\cite{recommender}, search~\cite{cancerkg}, and knowledge discovery~\cite{graphsage}. Modern web platforms naturally form large-scale \textit{text-attributed graphs} (TAGs), where nodes represent entities like users, products, or documents, and edges capture their structural or semantic relationships. Pre-training GFMs on such TAGs enables the extraction of transferable structural priors from massive online corpora while reducing downstream fine-tuning costs \cite{DBLP:journals/corr/abs-2310-11829,GraphAdapter,ren2024survey,liu2023graphprompt,sun2022gppt,allinone,graphgpt,graphcontrol}. A major line of progress lies in cross-modal alignment on TAGs, which leverages the semantic richness of textual attributes to enhance graph representation learning \cite{brannon2024congrat,wen2023augmenting,li2023grenade,jin2023patton,glem}. Inspired by the success of CLIP in vision–language learning \cite{clip,cleanclip,cleanerclip,robustclip,badclip2024liang}, recent methods such as ConGraT \cite{brannon2024congrat}, G2P2 \cite{wen2023augmenting}, and GraphCLIP \cite{zhu2025graphclip} adopt dual encoders—a GNN\cite{gcn,gat,graphsage,wo2024graph,cluster-gcn} for graph structures and a Transformer for text—trained with contrastive objectives \cite{oord2018infonce}. These approaches pull matched graph–text pairs closer while pushing apart mismatched ones, thereby transferring knowledge from pre-trained language models into graph encoders and supporting zero-/few-shot generalization across unseen domains and tasks. 

Despite these advances, existing contrastive approaches face two fundamental limitations that restrict their effectiveness on real-world TAGs. 
\textbf{Limitation 1: Overlooking many-to-many correspondences.} Most methods assume strict one-to-one alignments between nodes and texts, treating all unmatched pairs as negatives. They overlook the many-to-many relationships that are prevalent in practice: for example, a single scientific paper node may span multiple topics and thus be semantically related to several textual descriptions (see Figure~\ref{fig:motivation_illustrations}). 
\textbf{Limitation 2: Vulnerability to noisy supervision.} Existing approaches also rely heavily on the assumption of clean, perfectly matched pairs. When node–text associations are imperfect—due to annotation errors, ambiguous semantics, or weakly correlated modalities—they pull together mismatched pairs and severely mislead pre-training. Together, these limitations expose a core dilemma: many-to-many alignment enriches supervision but amplifies noise, while strict one-to-one alignment improves robustness but sacrifices semantic diversity. This dilemma underscores the dependence of current methods on data cleanliness, motivating the search for a more adaptive solution.

These two limitations highlight a fundamental trade-off that prevents a universal, static solution. On the one hand, incorporating many-to-many correspondences can capture richer semantic relations but also increases vulnerability to alignment noise, as misaligned groups may be incorrectly pulled together~\cite{gao2024softclip}. On the other hand, overly conservative robust losses reduce the impact of noisy supervision but risk discarding subtle yet meaningful signals when the data is clean~\cite{symnce,robustcl}. Recent advances in the vision–language community attempt to address each issue separately, but not both simultaneously. For example, soft-target approaches leverage intra-modal similarity to generate many-to-many signals~\cite{gao2024softclip}, enriching supervision at the cost of greater noise sensitivity. In contrast, robust loss designs~\cite{symnce,robustcl} enhance resilience to outliers but sacrifice semantic diversity. This gap raises a key research question:  \textit{Can we design a graph–text alignment objective that adapts dynamically between one-to-one and many-to-many relations, guided by the quality of pre-training data, without requiring manual intervention?}

To this end, we introduce \model, a dynamic quality-aware graph–text alignment framework. \model extends existing CLIP-style graph–text aligners with a hybrid objective that combines one-to-one correspondence with many-to-many alignment signals derived from intra-modal similarities. These two components are dynamically adjusted based on pre-training data quality, making best use of rich alignment signals while preventing many-to-many signals from being misled by noise. Concretely, to address \textbf{Limitation~1 (missing many-to-many relations)}, we construct soft alignment targets using fine-grained intra-modal similarities, allowing a graph node not only to align with its directly associated text, but also with other semantically related texts—and vice versa. A subgraph-level alignment loss further ensures consistent alignment at a higher structural level. To address \textbf{Limitation~2 (sensitivity to noisy supervision)}, we split the alignment objective into two components: (i) a standard contrastive loss enforcing strong one-to-one correspondences, and (ii) a soft alignment loss guided by intra-modal similarity and subgraph structure. These components are adaptively balanced by a dynamically estimated ratio $\theta$, which is derived from the alignment certainty within each mini-batch and reflects pre-training data quality. Intuitively, when intra-modal signals are consistent, \model emphasizes expressive many-to-many relations; when signals are uncertain or noisy, it downweights ambiguous many-to-many relations and focuses on high-confidence one-to-one alignments via adaptive filtering. In essence, \model adapts its strategy to data quality—promoting semantic richness on clean data and robustness under noise.  


Theoretically, our analysis provides formal guarantees for the stability and convergence of the dynamic mechanism. Comprehensive experiments on nine diverse TAG benchmarks validate our approach against 16 competing methods in various settings. The results show that ADAligner achieves strong and competitive performance, including zero-/few-shot node classification, link prediction, and cross-modal retrieval, in unsupervised and transfer scenarios, while maintaining strong robustness even under 30\% label noise. Furthermore, compared to leading multimodal baselines such as G2P2~\cite{wen2023augmenting} and GraphCLIP~\cite{zhu2025graphclip}, ADAligner accelerates pre-training by $2$–$3\times$ by replacing their costly strategies with its lightweight adaptive mechanism, establishing its effectiveness as a robust, efficient, and scalable framework.

\begin{figure*}[t]
    \centering
    \captionsetup{skip=3pt}
    \includegraphics[width=0.9\textwidth]{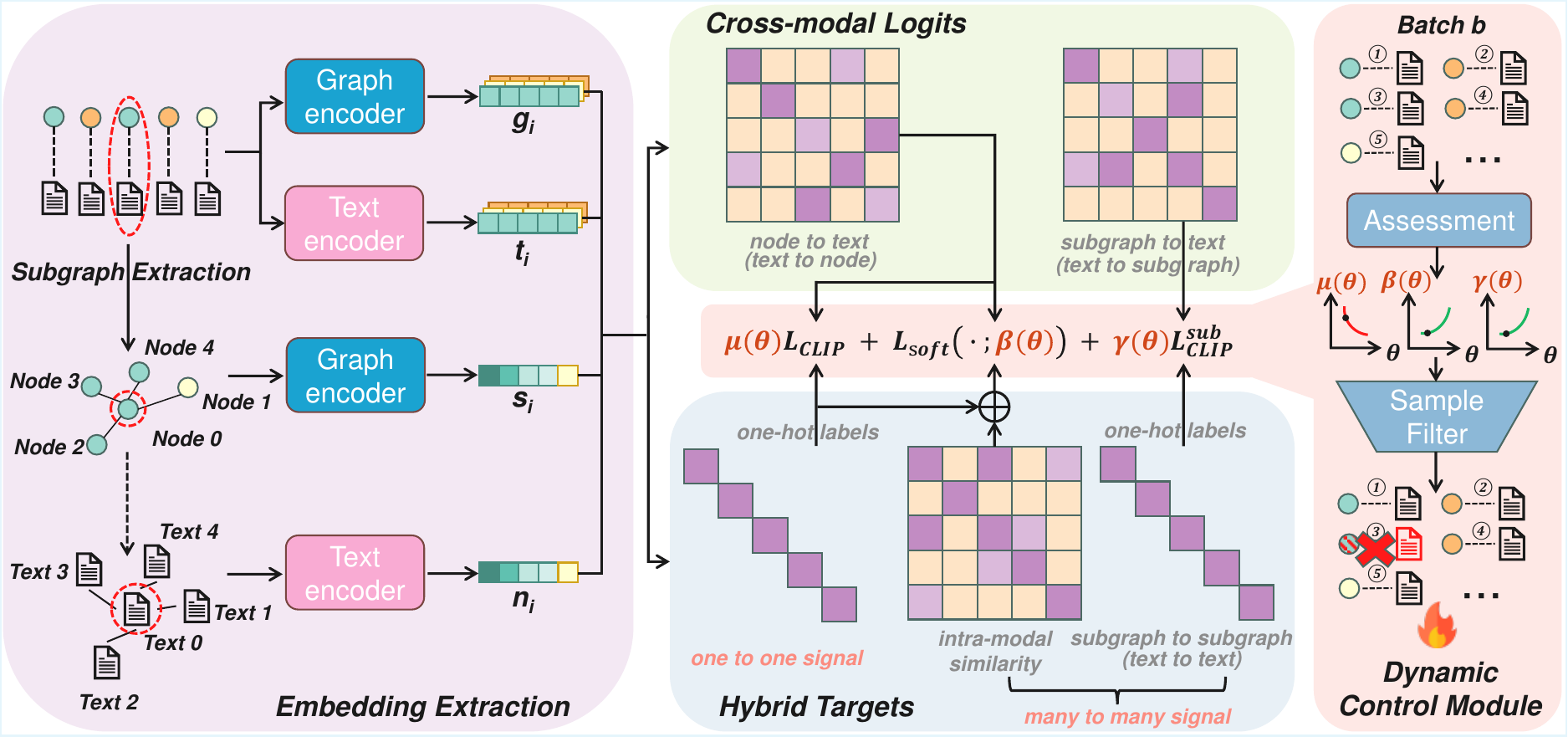} 
    \caption{Overview of the \model framework. Two modality encoders are aligned through a hybrid objective that combines one-to-one and many-to-many targets. The loss terms are automatically re-weighted by a data-quality–driven factor $\theta$.}
    \label{fig:framework_overview}
\end{figure*}

Our main contributions are summarized as follows:
\begin{itemize}[leftmargin = 8pt]
    \item We propose \model, a novel and principled graph-text alignment framework. Its core is a dynamic quality assessment mechanism that performs real-time evaluation of alignment confidence to adaptively balance expressive many-to-many objectives against robust one-to-one strategies while filtering out noisy pairs.
    
    \item We provide formal theoretical guarantees for our dynamic framework. Our analysis proves the stability of the adaptive controller and the convergence of the overall training objective, ensuring a reliable and principled learning process.
    
    \item We demonstrate through comprehensive experiments on diverse TAGs that ADAligner simultaneously achieves strong performance across multiple tasks, high robustness to noise, and a $2$–$3\times$ pre-training speedup over strong competitors.
\end{itemize}

\section{Preliminaries}
\label{sec:preliminaries}

\noindent \textbf{Notations.} 
We define a TAG as $\mathcal{G}=(\mathcal{V}, \mathcal{E}, \mathbf{X}, \mathcal{T})$, where $\mathcal{V}=\{v_1,\dots,v_N\}$ is the set of nodes, $\mathcal{E}$ the set of edges, $\mathbf{X} \in \mathbb{R}^{N \times d_f}$ the node features, and $\mathcal{T}=\{T_1,\dots,T_N\}$ the corresponding textual descriptions. Our goal is to learn a graph encoder $f_G(\cdot;\phi_G)$ and a text encoder $f_T(\cdot;\phi_T)$ that map nodes and texts into a shared $d$-dimensional embedding space. With a projection head $P_G$, we obtain
\[
\mathbf{g}_i = P_G(f_G(v_i; \phi_G)), \quad \mathbf{t}_i = f_T(T_i; \phi_T).
\]
All embeddings are $\ell_2$-normalized: $\hat{\mathbf{g}}_i = \mathbf{g}_i / \|\mathbf{g}_i\|_2$ and $\hat{\mathbf{t}}_i = \mathbf{t}_i / \|\mathbf{t}_i\|_2$.

\noindent \textbf{Standard One-to-One Contrastive Alignment.} 
The standard approach is the InfoNCE loss~\cite{oord2018infonce}, which enforces strict one-to-one correspondence within a mini-batch $\mathcal{B} = \{(v_i, T_i)\}_{i=1}^{N_B}$. Each paired $(v_i, T_i)$ is treated as positive, while all others are negatives. The symmetric contrastive loss is
\begin{equation}
\begin{aligned}
\mathcal{L}_{\text{CLIP}}(\mathcal{B}) = -\frac{1}{N_B} \sum_{i=1}^{N_B} \Biggl[ 
& \log \frac{\exp(\langle \hat{\mathbf{g}}_i, \hat{\mathbf{t}}_i \rangle / \tau)}{\sum_{j=1}^{N_B} \exp(\langle \hat{\mathbf{g}}_i, \hat{\mathbf{t}}_j \rangle / \tau)} \\
+ & \log \frac{\exp(\langle \hat{\mathbf{t}}_i, \hat{\mathbf{g}}_i \rangle / \tau)}{\sum_{j=1}^{N_B} \exp(\langle \hat{\mathbf{t}}_i, \hat{\mathbf{g}}_j \rangle / \tau)} 
\Biggr],
\end{aligned}
\label{eq:infonce_one_to_one}
\end{equation}
where $\tau$ is a learnable temperature parameter. This loss enforces one-to-one matching but ignores many-to-many correspondences and is vulnerable to noise—limitations we address in the next section.

\section{The \model Framework}
\label{sec:method}

\noindent \textbf{Overview}. Existing graph–text alignment methods often rely on overly strict one-to-one matching, which fails to capture many-to-many semantic correspondences and makes them highly sensitive to noisy supervision. To overcome these limitations, we propose \textbf{\model}, a dynamic quality-aware framework that jointly models richer many-to-many relations while remaining robust under noisy conditions. The key idea is to adaptively balance expressive many-to-many alignment and reliable one-to-one alignment based on the quality of the training data. The framework is illustrated in Figure~\ref{fig:framework_overview}.

\subsection{Modeling Diverse Graph-Text Relationships}
\label{sec:relationship_modeling}

The standard one-to-one contrastive loss (Eq.~\eqref{eq:infonce_one_to_one}) enforces rigid node–text correspondences that misalign with the complex semantics of real-world TAGs. While effective for clean, unambiguous pairs, it fails to capture the prevalent one-to-many or many-to-many relations—for instance, a scientific paper may relate to multiple topical descriptions, and communities of nodes may share overlapping semantics. Relying solely on $\mathcal{L}_{\text{CLIP}}$ thus discards valuable supervisory signals and limits the model’s ability to exploit the semantic richness of the graph. To overcome this, \model introduces a \textbf{many-to-many relationship learning objective} composed of two complementary components: a \textit{Soft Alignment Loss} that incorporates intra-modal similarity into cross-modal targets, and a \textit{Subgraph–Text Alignment Loss} that enforces neighborhood-level consistency.

\noindent \textbf{Soft Alignment Loss.}
To move beyond hard one-hot supervision, \model constructs soft alignment targets that incorporate intra-modal self-similarity. 
The key intuition is that the way a sample relates to its neighbors within the same modality provides valuable cues for determining its cross-modal correspondences. 

Concretely, for each sample $i$ in a batch $\mathcal{B}$, the soft target distribution is obtained by interpolating its one-hot target $\mathbf{y}_i$ (where $y_{ii}=1$ and $y_{ij}=0$ for $j \neq i$) with similarity-based probabilities. 
The graph-guided and text-guided targets are defined as:
\begin{align}
\tilde{\mathbf{p}}_{gg,i} &= (1-\beta)\mathbf{y}_i + \beta \cdot \text{softmax}(\mathbf{S}_{gg,i,:} / \tau), \label{eq:soft_target_gg_revised} \\
\tilde{\mathbf{p}}_{tt,i} &= (1-\beta)\mathbf{y}_i + \beta \cdot \text{softmax}(\mathbf{S}_{tt,i,:} / \tau), \label{eq:soft_target_tt_revised}
\end{align}
where $\mathbf{S}_{gg} \in \mathbb{R}^{N_B \times N_B}$ is the intra-graph similarity matrix computed from $\{\hat{\mathbf{g}}_k\}$, $\mathbf{S}_{tt}$ is the intra-text similarity matrix from $\{\hat{\mathbf{t}}_k\}$, $\beta \in [0,1]$ is a mixing coefficient, and $\tau$ is the temperature. The rows of $\mathbf{S}_{gg}$ and $\mathbf{S}_{tt}$ are normalized by softmax to form probability distributions.

Given these soft targets, the predicted cross-modal distributions are obtained from the graph-to-text similarity matrix $\mathbf{S}_{gt}$ and the text-to-graph similarity matrix $\mathbf{S}_{tg}$, respectively. 
The Soft Alignment Loss minimizes the symmetric KL divergence ($D_{\text{KL}}^{sym}(p\|q) = \frac{1}{2}(D_{\text{KL}}(p\|q) + D_{\text{KL}}(q\|p))$). The loss is thus:
\begin{equation}
\begin{aligned}
\mathcal{L}_{\text{soft}}(\mathcal{B}; \beta) 
= \tfrac{1}{2N_B} \sum_{i=1}^{N_B} \Big[ &
D_{\text{KL}}^{\text{sym}}\!\big(\tilde{\mathbf{p}}_{gg,i} \,\|\, \text{softmax}(\mathbf{S}_{gt,i,:}/\tau)\big) \\
&+ D_{\text{KL}}^{\text{sym}}\!\big(\tilde{\mathbf{p}}_{tt,i} \,\|\, \text{softmax}(\mathbf{S}_{tg,i,:}/\tau)\big) \Big].
\end{aligned}
\end{equation}
This loss aligns nodes with both paired and semantically related texts, capturing fine-grained many-to-many relations.

\noindent \textbf{Subgraph Structure–Text Alignment Loss.}
Beyond individual nodes, local neighborhoods encode important structural context. To capture this, \model builds a \textit{neighborhood structural embedding} $\mathbf{g}_{nb,i}$ and an \textit{aggregated neighborhood text embedding} $\mathbf{t}_{nb,i}$ for each node $v_i$ by averaging the embeddings of up to five randomly sampled 1-hop neighbors. These context-enriched embeddings are then aligned via the standard contrastive loss, denoted as $\mathcal{L}_{\text{CLIP}}^{sub}$:
\begin{equation}
\mathcal{L}_{\text{CLIP}}^{sub} =
\mathcal{L}_{\text{CLIP}}(\mathcal{B}; \{\mathbf{g}_{nb,k}\}, \{\mathbf{t}_{nb,k}\}).
\end{equation}
This encourages neighborhood-level structural–textual consistency, enabling the model to capture broader semantic relations that cannot be explained by single-node alignments.

\subsection{Dynamic Quality-Aware Adaptation}
\label{sec:dynamic_adaptation_v2}
While the many-to-many objectives in Section~\ref{sec:relationship_modeling} enrich supervision, they are also more vulnerable to spurious correlations in noisy data. To balance expressiveness with robustness, \model introduces a \textbf{dynamic quality-aware adaptation mechanism} that operates at the mini-batch level. The key intuition is that reliable alignments exhibit a clear similarity margin between matched and unmatched pairs, whereas noisy or ambiguous pairs blur this margin. We therefore use the margin as a proxy for alignment confidence~\cite{wang2019gpw,robinson2021contrastivelearninghardnegative}. Based on this assessment, \model adjusts its training in two ways: (i) re-weighting the relative contributions of one-to-one and many-to-many objectives, and (ii) filtering out low-confidence pairs. Together, these mechanisms allow \model to automatically adapt its learning strategy to the quality of the data. A theoretical analysis of stability and convergence is provided in Section~\ref{subsec:theoretical_analysis}.

\subsubsection{Batch Quality Assessment and Control Factor $\theta$} 
To enable dynamic adaptation, \model first estimates the quality of each mini-batch $\mathcal{B}$. 
For every sample $i$, an individual quality score is computed as
\begin{equation}
M_i = S_{ii} - \mathbb{E}_{j \neq i}[S_{ij}],
\end{equation}
where $S_{ij} = \langle \hat{\mathbf{g}}_i, \hat{\mathbf{t}}_j \rangle$ denotes the cross-modal similarity between graph embedding $\hat{\mathbf{g}}_i$ and text embedding $\hat{\mathbf{t}}_j$. 
Intuitively, $M_i$ measures how well sample $i$ distinguishes its correct partner from negatives. 
The batch-level quality is then given by the average score:
\begin{equation}
M_{\mathcal{B}} = \mathbb{E}_{i \in \mathcal{B}}[M_i].
\end{equation}

To ensure stability, \model maintains exponential moving averages (EMAs) of both the historical mean ($M_0$) and variance ($\sigma_0^2$) of the $M_i$ values, updated with momentum $m$:
\begin{align}
M_0^{(t)} &\leftarrow m M_0^{(t-1)} + (1 - m) M_{\mathcal{B}}^{(t)}, \label{eq:ema_mean_theta_focus} \\
(\sigma_0^2)^{(t)} &\leftarrow m (\sigma_0^2)^{(t-1)} + (1 - m)\,\text{Var}_{i \in \mathcal{B}^{(t)}}[M_i]. \label{eq:ema_var_theta_focus}
\end{align}

Finally, a control factor $\theta$ is derived by comparing the current batch quality to the historical mean:
\begin{equation}
\theta = \theta_{0} + \alpha\,(M_{\mathcal{B}} - M_0),
\label{eq:theta_definition}
\end{equation}
where $\theta_{0}$ is the base value and $\alpha$ controls sensitivity to deviations.
Larger $\theta$ implies cleaner data; smaller indicates noise.

\subsubsection{Orchestration by Control Factor $\theta$: Objective Adaptation and Filtering} 
\label{sec:theta_orchestration}
The  $\theta$, derived from batch quality assessment, plays a central role in \model's adaptive behavior by dictating both the balance of learning objectives and selected samples.

\noindent \textbf{Adaptive Learning Objective via $\theta$-Modulated Parameters.}
$\theta$ dynamically adjusts the emphasis between learning many-to-many relationships and enforcing robust one-to-one alignment by modulating the parameters of the loss components. 
For the \textit{Many-to-Many Relationship Learning Objective}, which comprises $\mathcal{L}_{soft}$ and $\mathcal{L}_{CLIP}^{sub}$, their respective controlling parameters are $\beta(\theta)$ and $\gamma(\theta)$. 
For the \textit{One-to-One Relationship Learning Objective}, which consists of the standard contrastive loss $\mathcal{L}_{CLIP}$, its weight is $\mu(\theta)$. 
Formally:
\begin{equation}
\underbrace{\beta(\theta) = \theta \cdot \beta_{0},~~~~~\gamma(\theta) = \theta \cdot \gamma_{0}}_{\text{Many-to-Many Alignment Objective}},~~~~~
\underbrace{\mu(\theta) = \tfrac{\mu_{0}}{1 + \theta},}_{\text{One-to-One Alignment Objective}}
\label{eq:mu_orchestrated}
\end{equation}
where $\beta_0, \gamma_0, \mu_0$ are base hyperparameters. 
To ensure stability and avoid degenerate solutions, these dynamically modulated coefficients are constrained within bounded ranges: $\beta \in [0.1, 0.9]$, $\mu \in [0.1, 1.9]$, and $\gamma \in [0.1, 1.9]$, enforced via clamping in implementation.
This guarantees that no single objective dominates training excessively, while still allowing sufficient flexibility for adaptation. 
As a result, $\beta(\theta)$ and $\gamma(\theta)$ increase with $\theta$, strengthening many-to-many objectives on cleaner batches; conversely, $\mu(\theta)$ increases as $\theta$ decreases, prioritizing robust one-to-one alignment on noisier batches.

\noindent \textbf{Dynamic Sample Filtering via $\theta$.} 
In addition to modulating loss parameters, $\theta$ also controls the intensity of sample filtering. 
A straightforward strategy is to deterministically retain the top-$k$ samples with the highest quality scores ($M_i$). 
However, this greedy, 'winner-take-all' approach, while capturing strong signals, risks overfitting the model to a few 'perfect' examples. 
It largely ignores the global data distribution by discarding a wealth of 'good-enough' samples that are crucial for generalization. 
Therefore, a more robust mechanism is needed---one that can not only preserve high-quality outliers but also maintain consistency with the global data distribution.

To this end, we design a stochastic filtering mechanism, inspired by the principles of $\beta$-DPO~\cite{wu2024betadpo}, that elegantly resolves this trade-off. 
For each sample $i$ in batch $\mathcal{B}$, we first compute its individual quality score $M_i = S_{ii} - \mathbb{E}_{j \neq i}[S_{ij}]$. 
We then use a Gaussian function centered at the historical mean quality $M_0$ to assign a sampling weight $w_i$:
\begin{equation} \label{eq:weight}
w_i = \exp\!\left(-\tfrac{1}{2}\bigl(\tfrac{M_i - M_0}{\sigma}\bigr)^2\right).
\end{equation}
To convert these weights into a valid probability distribution for sampling, we then normalize them across all samples in the batch $\mathcal{B}$:
\begin{equation} \label{eq:prob}
p_i = \frac{w_i}{\sum_{j \in \mathcal{B}} w_j}.
\end{equation}
The resulting weighting scheme naturally assigns a higher probability to samples consistent with the historical mean. 
Then, instead of deterministically picking the most probable samples, we perform multinomial sampling from the distribution $\{p_i\}$.
The number of samples to retain, $N_{\text{keep}}$, is determined by a dynamic retention ratio, $\rho(\theta)$, which adjusts based on the deviation of the current batch quality $\theta$ from a baseline value $\theta_0$:
\begin{equation} \label{eq:ratio}
\rho(\theta) = \rho_0 + \lambda (\theta - \theta_0),
\end{equation}
where $\rho_0$ is the base retention ratio and $\lambda$ is a sensitivity hyperparameter. For stability, the final ratio is clipped to a valid range (e.g., $[\rho_{\min}, 1.0]$). The number of samples to retain is then calculated as:
\begin{equation} \label{eq:nkeep}
N_{\text{keep}} = \max(1, \lfloor \rho(\theta) \cdot |\mathcal{B}| \rfloor).
\end{equation}
This stochastic process ensures that while consistent samples are favored, high-quality outliers still have a chance of being selected. 
As the historical mean $M_0$ progressively increases during training, the Gaussian weighting naturally shifts toward higher-quality samples, inducing a curriculum-like effect that emphasizes robustness in early stages and precision in later stages. 
Consequently, \model can maintain robustness against noise while avoiding the loss of informative but rare signals. 
As shown in Section~\ref{subsec:filter}, the precision of our dynamic filter at discarding mismatched pairs progressively improves during training (Figure~\ref{fig:filter}).

\subsection{Total Adaptive Loss Objective}
\label{sec:total_loss_final} 
\model's final training objective integrates these dynamically controlled components. For each batch $\mathcal{B}$, the control factor $\theta$ dictates the sample filtering process (resulting in $\mathcal{B}^f$) and dynamically sets the parameters $\beta(\theta)$ for $\mathcal{L}_{soft}$, and the weights $\gamma(\theta)$ for $\mathcal{L}_{CLIP}^{sub}$ and $\mu(\theta)$ for $\mathcal{L}_{CLIP}$. The overall loss is:
\begin{equation}
    \mathcal{L}_{\text{total}} = \mathbb{E}_{\mathcal{B}} \Biggl[  \mathcal{L}_{\text{soft}}(\mathcal{B}^f; \beta(\theta)) + \gamma(\theta) \mathcal{L}_{\text{CLIP}}^{\text{sub}}(\mathcal{B}^f) 
     + \mu(\theta) \mathcal{L}_{\text{CLIP}}(\mathcal{B}^f) \Biggr].
\label{eq:loss_total_final}
\end{equation}
This unified objective thereby enables \model to learn expressive and robust representations by adaptively balancing the emphasis on many-to-many relationship modeling and one-to-one alignment based on real-time data quality.

\subsection{Theoretical Analysis}
\label{subsec:theoretical_analysis}

We now analyze the stability and convergence of our dynamic mechanism. 
The total loss $\mathcal{L}_{\text{total}}(\Phi, \theta)$ can be viewed as a time-varying objective, where the variation is induced by the adaptive controller $\theta_t$. 
Our analysis proceeds in three steps: we first establish the boundedness of all key signals, then show the stability of the controller $\theta_t$, and finally prove that the overall training loss converges. 
Formal proofs are deferred to Appendix~\ref{sec:theoretical_analysis}.

\begin{lemma}[Signal Boundedness]
\label{lem:boundedness}
For L2-normalized embeddings and batch size $N_B \ge 2$, the batch quality score $M_{\mathcal{B}}$, the controller $\theta$, and the dynamic weights $\beta(\theta),\gamma(\theta),\mu(\theta)$ are all bounded within fixed ranges.
\end{lemma}

\noindent\textbf{Remarks.} Since both $M_{\mathcal{B}}$ and its EMA $M_0$ are bounded, their difference cannot cause $\theta$ to diverge. The functional forms of $\beta(\theta),\gamma(\theta),\mu(\theta)$ also avoid singularities, ensuring all loss terms remain well-defined during training.

\medskip
Having established that all driving signals remain bounded, we can next analyze the dynamical behavior of the controller itself. In particular, we examine whether the iterative update of $\theta$ remains stable over time.

\begin{lemma}[Dynamic Stability]
\label{lem:stability}
The controller is updated according to
\[
\theta_{t+1} \;=\; \Pi_{[\theta_{\min},\theta_{\max}]}\!\Big(\theta_t 
+ \alpha\big(M_{\mathcal{B}}(\Phi_t) - M_{0,t}\big)\Big),
\]
where $M_{0,t}$ is the EMA of the batch quality score $M_{\mathcal{B}}(\Phi_t)$.  
Under smoothness and negative-drift assumptions, and with sufficiently small step sizes $\eta,\alpha$, the controller $\theta_t$ is stable in expectation and converges to a neighborhood of an equilibrium point $\theta^*$.
\end{lemma}

\noindent\textbf{Remarks.} The negative-drift assumption ensures that $\theta$ self-corrects. When $\theta$ is too high, the model emphasizes expressive many-to-many objectives, which increase $\mathcal{L}$ and lower $M_{\mathcal{B}}$, pulling $\theta$ downward. Conversely, when $\theta$ is too low, the model falls back to robust one-to-one alignment, which improves $M_{\mathcal{B}}$, pushing $\theta$ upward. Thus $\theta_t$ oscillates around $\theta^*$ without divergence. This theoretical stability is further supported by empirical results (Appendix~\ref{app:theta_stability}), where $\theta$ consistently converges across different data conditions.

\medskip
With both boundedness and stability in place, we are now equipped to analyze the behavior of the overall training objective. The final step is to show that the total loss converges under standard assumptions on the loss components and gradient updates.

\begin{theorem}[Convergence of Total Loss]
\label{thm:convergence_main}
Suppose each component loss is differentiable and lower-bounded, $\mathcal{L}_{\text{total}}(\cdot,\theta)$ is $L_\Phi$-smooth in $\Phi$, $\mathcal{L}_{\text{total}}(\Phi,\cdot)$ is $L_\theta$-Lipschitz in $\theta$, the stochastic gradient is unbiased with bounded variance, and the controller $\theta_t$ evolves on a slower time scale than the model parameters $\Phi$. Then gradient descent ensures
\[
\liminf_{t \to \infty} \mathbb{E}[\|\nabla_\Phi \mathcal{L}_{\text{total}}(\Phi_t, \theta_t)\|] = 0.
\]
That is, the total loss converges in expectation to a bounded neighborhood around a stationary point $(\Phi^*, \theta^*)$, 
where the neighborhood radius depends on the variance of the stochastic gradient and the adaptation rate of $\theta_t$.
\end{theorem}

\noindent\textbf{Remarks.}
The two-time-scale design guarantees that $\theta_t$ can be treated as quasi-static relative to $\Phi_t$, 
allowing standard convergence arguments for stochastic gradient descent to hold up to small perturbations. 
These perturbations are bounded by the EMA smoothing of $M_0$ and the limited step size of $\theta$, 
ensuring convergence to a stable equilibrium rather than divergence.

\begin{table*}[t]
\centering
\captionsetup{skip=1pt}
\caption{Unsupervised node classification ACC on clean datasets and under 30\% noise. 
The highest results are highlighted in \textbf{bold} and the second-best results are \underline{underlined}. 
F1-score is reported in Table~\ref{tab:unsupervised_f1_combined}.}
\label{tab:unsupervised_acc_combined}
\resizebox{1.0\linewidth}{!}{%
  \setlength{\tabcolsep}{4pt}
  \footnotesize
  \begin{tabular}{@{}l*{12}{c}@{}}
  \toprule
  \multirow{2}{*}{Methods} 
  & \multicolumn{2}{c}{Pubmed} & \multicolumn{2}{c}{Cora} & \multicolumn{2}{c}{Citeseer} & \multicolumn{2}{c}{WikiCS} & \multicolumn{2}{c}{Reddit} & \multicolumn{2}{c}{Instagram} \\
  \cmidrule(lr){2-3} \cmidrule(lr){4-5} \cmidrule(lr){6-7} \cmidrule(lr){8-9} \cmidrule(lr){10-11} \cmidrule(lr){12-13}
   & 0\% & 30\% & 0\% & 30\% & 0\% & 30\% & 0\% & 30\% & 0\% & 30\% & 0\% & 30\% \\
  \midrule
  GCN\cite{gcn}   & ${62.97}_{\pm 2.17}$ & -- & ${69.52}_{\pm 2.28}$ & -- & ${59.40}_{\pm 5.31}$ & -- & ${43.84}_{\pm 2.89}$ & -- & ${52.97}_{\pm 0.44}$ & -- & ${62.90}_{\pm 1.34}$ & -- \\
  GAT\cite{gat}   & ${63.62}_{\pm 0.66}$ & -- & ${68.78}_{\pm 4.40}$ & -- & ${55.42}_{\pm 3.19}$ & -- & ${46.72}_{\pm 3.92}$ & -- & ${55.54}_{\pm 0.60}$ & -- & $\underline{{63.41}}_{\pm 1.28}$ & -- \\
  DGI\cite{dgi}   & ${66.29}_{\pm 1.51}$ & -- & ${71.33}_{\pm 3.34}$ & -- & ${37.30}_{\pm 3.97}$ & -- & ${39.86}_{\pm 1.57}$ & -- & $\mathbf{64.70}_{\pm 0.50}$ & -- & ${54.98}_{\pm 4.83}$ & -- \\
  GRACE\cite{grace} & ${63.71}_{\pm 2.05}$ & -- & $\underline{{72.13}}_{\pm 1.52}$ & -- & ${65.14}_{\pm 2.48}$ & -- & ${51.20}_{\pm 0.92}$ & -- & ${53.93}_{\pm 0.85}$ & -- & ${62.50}_{\pm 1.82}$ & -- \\
  GraphCL\cite{gcl} & ${62.48}_{\pm 1.95}$ & -- & ${68.08}_{\pm 4.19}$ & -- & ${55.67}_{\pm 6.18}$ & -- & ${46.78}_{\pm 1.74}$ & -- & ${52.88}_{\pm 0.89}$ & -- & ${63.01}_{\pm 1.45}$ & -- \\ \midrule
  BERT\cite{devlin2019bert}    & ${47.04}_{\pm 0.40}$ & -- & ${41.29}_{\pm 2.20}$ & -- & ${45.61}_{\pm 1.68}$ & -- & ${36.02}_{\pm 0.51}$ & -- & ${51.40}_{\pm 1.30}$ & -- & ${57.39}_{\pm 0.26}$ & -- \\
  SBERT\cite{reimers2019sbert}   & ${60.04}_{\pm 1.12}$ & -- & ${60.18}_{\pm 2.80}$ & -- & ${64.73}_{\pm 1.19}$ & -- & ${57.72}_{\pm 0.04}$ & -- & ${52.28}_{\pm 0.67}$ & -- & ${39.92}_{\pm 1.10}$ & -- \\
  RoBERTa\cite{liu2019roberta} & ${37.90}_{\pm 0.55}$ & -- & ${41.59}_{\pm 1.06}$ & -- & ${39.40}_{\pm 2.76}$ & -- & ${37.55}_{\pm 1.18}$ & -- & ${51.69}_{\pm 1.00}$ & -- & ${51.33}_{\pm 0.91}$ & -- \\
  DeBERTa\cite{he2020deberta} & ${47.03}_{\pm 1.37}$ & -- & ${40.37}_{\pm 1.15}$ & -- & ${36.83}_{\pm 3.23}$ & -- & ${34.29}_{\pm 1.05}$ & -- & ${50.84}_{\pm 0.38}$ & -- & ${58.54}_{\pm 5.46}$ & -- \\
  Qwen3-0.6B\cite{yang2025qwen3technicalreport} & ${46.55}_{\pm 5.31}$ & -- & ${46.13}_{\pm 2.30}$ & -- & ${43.54}_{\pm 3.19}$ & -- & ${39.48}_{\pm 3.67}$ & -- & ${52.00}_{\pm 0.80}$ & -- & ${54.33}_{\pm 1.27}$ & -- \\ \midrule
  G2P2\cite{wen2023augmenting}    & ${52.19}_{\pm 9.63}$ & ${47.35}_{\pm 7.98}$ & ${47.05}_{\pm 6.44}$ & ${37.64}_{\pm 8.67}$ & ${22.13}_{\pm 3.80}$ & ${19.91}_{\pm 2.93}$ & ${15.74}_{\pm 8.04}$ & ${21.17}_{\pm 10.1}$ & ${49.00}_{\pm 1.91}$ & ${49.92}_{\pm 0.60}$ & ${51.00}_{\pm 8.04}$ & ${52.09}_{\pm 9.04}$ \\
  G2P2*(2-shots)\cite{wen2023augmenting} & ${60.32}_{\pm 4.94}$ & ${50.41}_{\pm 4.27}$ & ${48.82}_{\pm 3.52}$ & ${42.29}_{\pm 3.57}$ & ${24.14}_{\pm 3.64}$ & ${21.16}_{\pm 2.84}$ & ${56.11}_{\pm 3.46}$ & ${53.39}_{\pm 2.74}$ & ${51.10}_{\pm 3.34}$ & ${50.42}_{\pm 0.84}$ & ${53.06}_{\pm 3.77}$ & ${53.42}_{\pm 3.57}$ \\
  G2P2*(8-shots)\cite{wen2023augmenting} & ${70.10}_{\pm 2.40}$ & ${59.90}_{\pm 3.05}$ & ${55.57}_{\pm 3.66}$ & ${50.44}_{\pm 2.16}$ & ${28.81}_{\pm 2.28}$ & ${24.83}_{\pm 1.85}$ & ${61.85}_{\pm 1.04}$ & ${57.75}_{\pm 1.71}$ & ${50.48}_{\pm 3.97}$ & ${51.65}_{\pm 2.56}$ & ${51.83}_{\pm 3.77}$ & ${52.68}_{\pm 3.87}$ \\
  ConGraT\cite{brannon2024congrat} & ${56.40}_{\pm 4.63}$ & ${54.12}_{\pm 5.15}$ & ${21.46}_{\pm 3.11}$ & ${17.22}_{\pm 1.73}$ & ${21.93}_{\pm 2.70}$ & ${18.94}_{\pm 1.55}$ & ${15.29}_{\pm 3.21}$ & ${10.35}_{\pm 5.81}$ & ${53.70}_{\pm 1.26}$ & ${54.74}_{\pm 1.36}$ & ${50.85}_{\pm 4.06}$ & ${51.96}_{\pm 2.12}$ \\ \midrule
  \rowcolor{mygray}
  \model & $\mathbf{77.05}_{\pm 2.16}$ & $\mathbf{75.67}_{\pm 1.48}$ & $\mathbf{72.34}_{\pm 2.21}$ & $\mathbf{70.30}_{\pm 2.02}$ & $\mathbf{71.51}_{\pm 1.17}$ & $\mathbf{67.88}_{\pm 1.62}$ & $\mathbf{63.78}_{\pm 1.12}$ & $\mathbf{62.02}_{\pm 2.01}$ & $\underline{56.02}_{\pm 1.31}$ & $\mathbf{55.15}_{\pm 1.12}$ & $\mathbf{63.54}_{\pm 1.29}$ & $\mathbf{63.04}_{\pm 1.18}$ \\
  \model-CLIP & ${73.39}_{\pm 3.17}$ & ${70.37}_{\pm 2.45}$ & ${68.40}_{\pm 2.07}$ & ${65.27}_{\pm 1.67}$ & ${67.74}_{\pm 2.66}$ & ${61.35}_{\pm 5.60}$ & ${58.80}_{\pm 3.12}$ & ${57.63}_{\pm 2.79}$ & ${46.91}_{\pm 1.77}$ & ${45.45}_{\pm 1.46}$ & ${42.41}_{\pm 1.55}$ & ${42.53}_{\pm 2.24}$ \\
  w/o Assessment & ${73.76}_{\pm 3.15}$ & ${72.89}_{\pm 3.17}$ & ${69.08}_{\pm 2.21}$ & ${68.12}_{\pm 2.99}$ & ${68.87}_{\pm 1.78}$ & ${64.98}_{\pm 2.45}$ & ${61.94}_{\pm 1.44}$ & ${58.65}_{\pm 2.25}$ & ${54.71}_{\pm 3.09}$ & ${54.03}_{\pm 2.31}$ & ${59.52}_{\pm 1.84}$ & ${58.91}_{\pm 1.62}$ \\
  w/o Filter & $\underline{75.98}_{\pm 3.03}$ & $\underline{72.95}_{\pm 1.79}$ & ${70.25}_{\pm 3.32}$ & $\underline{68.45}_{\pm 1.38}$ & $\underline{69.32}_{\pm 1.25}$ & $\underline{65.26}_{\pm 2.14}$ & $\underline{62.53}_{\pm 1.73}$ & $\underline{60.81}_{\pm 1.77}$ & ${53.91}_{\pm 1.92}$ & $\underline{54.80}_{\pm 1.17}$ & ${60.09}_{\pm 2.18}$ & $\underline{59.47}_{\pm 2.55}$ \\ 
  \bottomrule
  \end{tabular}%
}
\end{table*}

\section{Experiments}
\label{sec:experiments}

In this section, we investigate the following research questions: 

\noindent \textbf{RQ1:} Can \model learn high-quality graph–text alignments in the unsupervised setting?  

\noindent \textbf{RQ2:} How well does \model generalize to unseen graphs/tasks in the transfer setting?  

\noindent \textbf{RQ3:} How do the dynamic controller $\theta$ and filtering mechanism behave in practice, and do they reliably ensure stability and robustness under noisy conditions? 

\noindent \textbf{RQ4:} How sensitive is \model to hyperparameter choices?  

\subsection{Experimental Setup}
\label{sec:setup}

We first describe the experimental setup before presenting results for each research question. We evaluate under both \textit{clean} and \textit{noisy} alignment scenarios, where $30\%$ of training text attributes are randomly swapped across classes to simulate annotation noise. 
Performance is assessed on three key downstream tasks—zero/few-shot node classification, link prediction, and cross-modal retrieval—averaged over 10 independent runs. Unless otherwise specified, all experiments share the same training configuration, with full details given in Appendix~\ref{app:training_hyper}.

\noindent \textbf{Datasets.} 
We consider a diverse set of TAGs spanning citation, social, and e-commerce domains. 
In the \textit{unsupervised setting}, six datasets are used: Cora, Citeseer, Pubmed, WikiCS, Reddit, and Instagram~\cite{li2024glbench}. 
In the \textit{transfer setting}, following GraphCLIP~\cite{zhu2025graphclip}, we pre-train on Pubmed and Reddit and transfer to seven unseen targets: Cora, Citeseer, WikiCS, Instagram, Ele-Photo, Ele-Computers, and Ele-History. 
All evaluations are performed on clean test sets; see Appendix~\ref{app:datasets_in_setup} for more details.

\noindent \textbf{Baselines.} 
We compare with three families of methods:  
(1)~\textit{Graph SSL models} that learn node embeddings without text supervision;  
(2)~\textit{Text-only encoders} that directly embed node attributes;  
(3)~\textit{Graph–text aligners}, including recent multimodal approaches.
In the transfer setting, we further include ZeroG~\cite{li2024zerog} as a recent benchmark. All implementations follow official settings. A full baseline list with detailed configurations is provided in Appendix~\ref{app:baseline_details_in_setup}.

\begin{figure*}[htbp]
    \centering
    \captionsetup{skip=2pt}
    \begin{minipage}[b]{0.48\textwidth}
        \centering
        \begin{subfigure}[b]{0.49\textwidth}
            \centering
            \includegraphics[width=\linewidth]{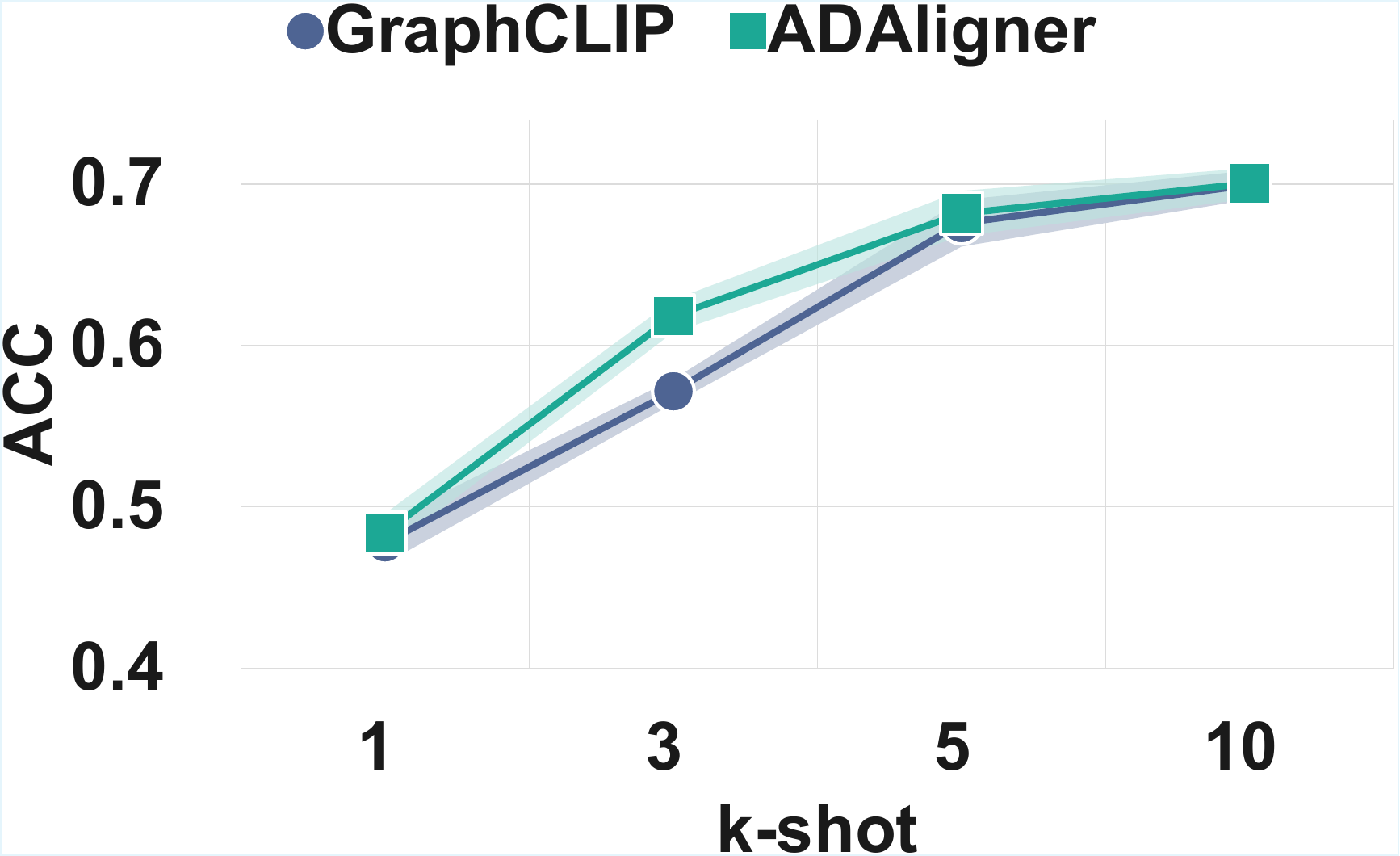}
            \caption{WikiCS (Clean)}
            \label{fig:few_shot_clean}
        \end{subfigure}
        \hfill %
        \begin{subfigure}[b]{0.49\textwidth}
            \centering
            \includegraphics[width=\linewidth]{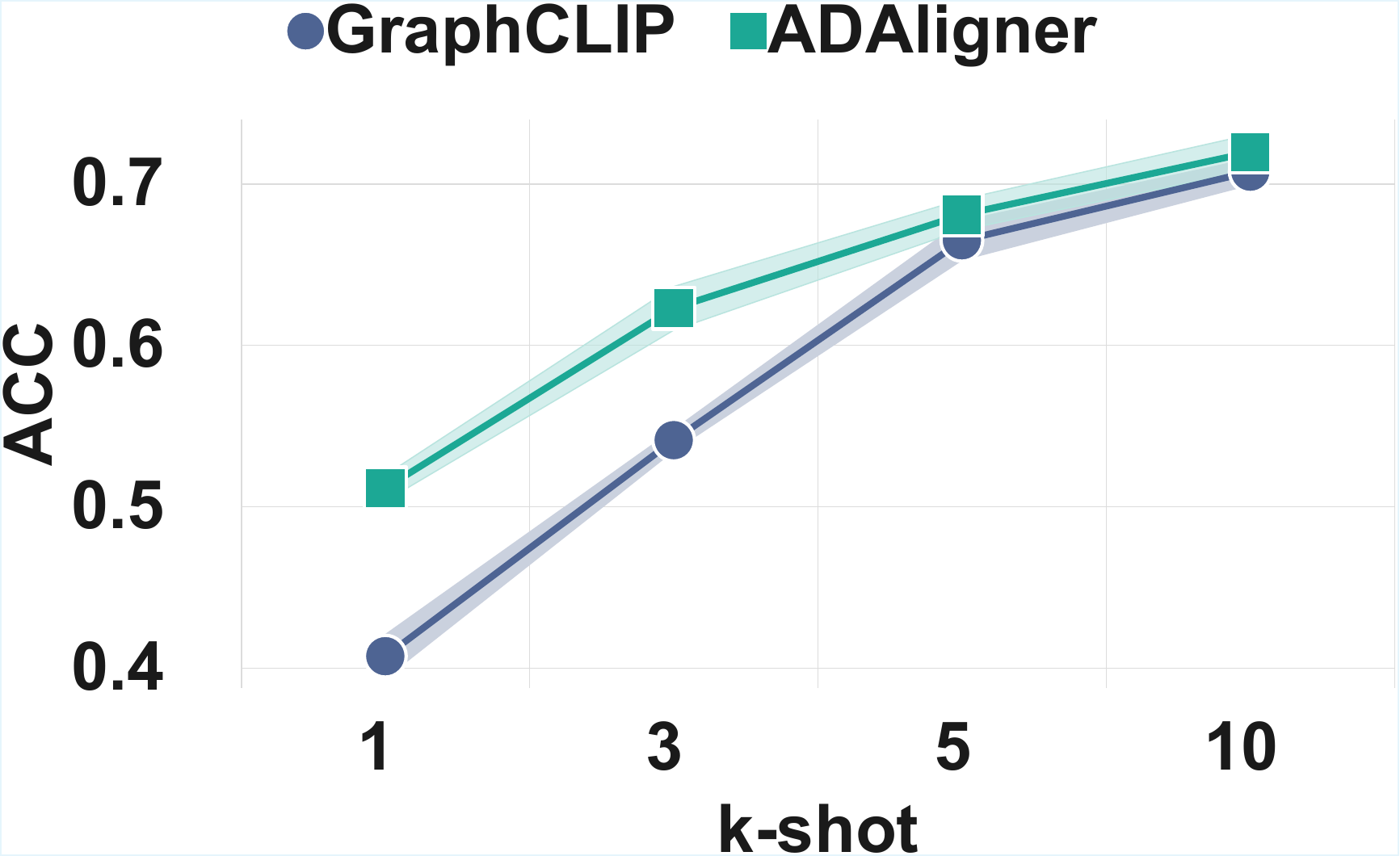}
            \caption{WikiCS (Noisy)}
            \label{fig:few_shot_noisy}
        \end{subfigure}
        \caption{Few-shot node classification ACC on WikiCS.}
        \label{fig:few_shot_results_wikics}
    \end{minipage}\hfill
    \begin{minipage}[b]{0.24\textwidth}
        \centering
        \includegraphics[width=\linewidth]{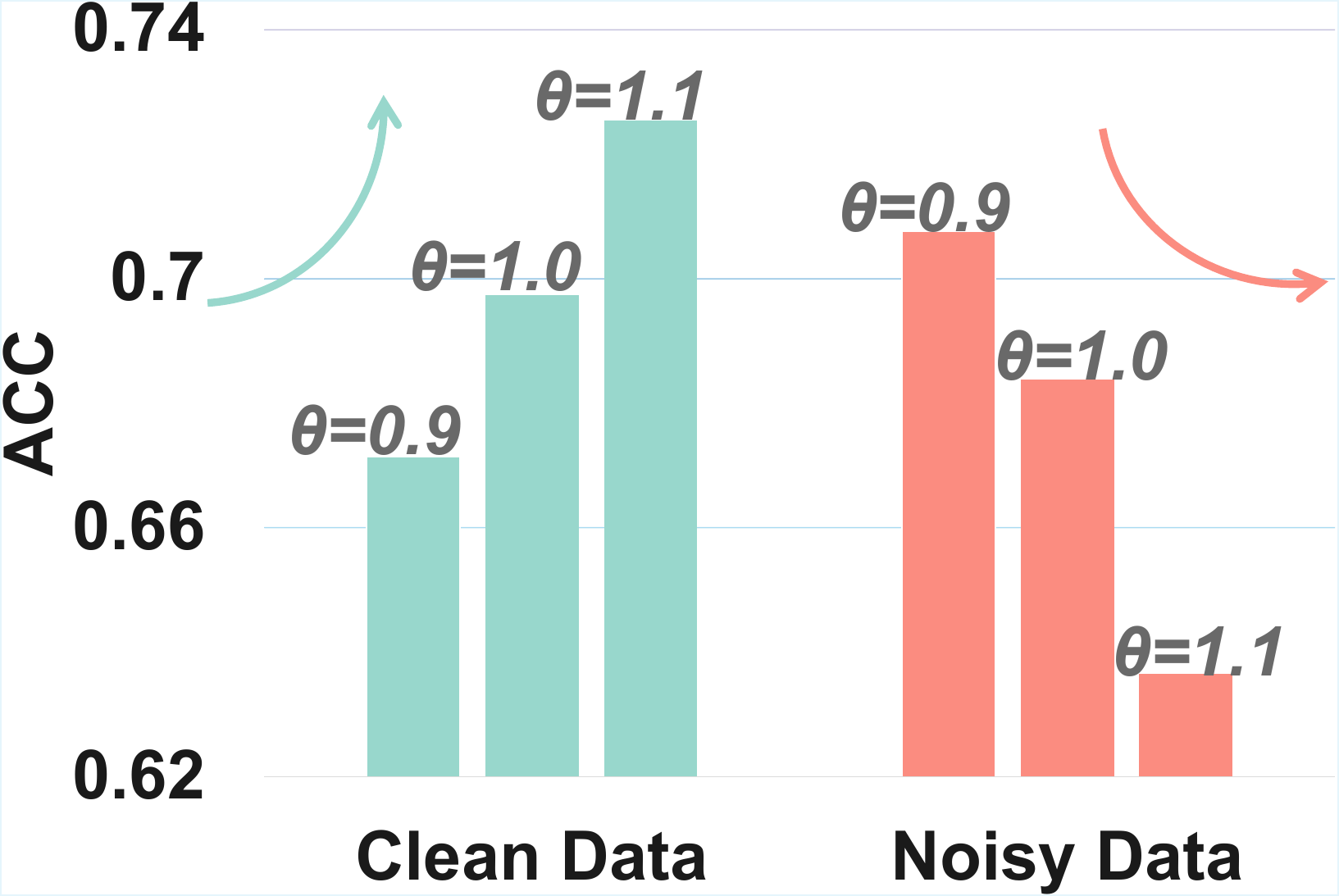}
        \caption{Fixed $\theta$ analysis under clean and noisy.}
        \label{fig:fixed_theta_performance_en}
    \end{minipage}\hfill
    \begin{minipage}[b]{0.24\textwidth}
        \centering
        \includegraphics[width=\linewidth]{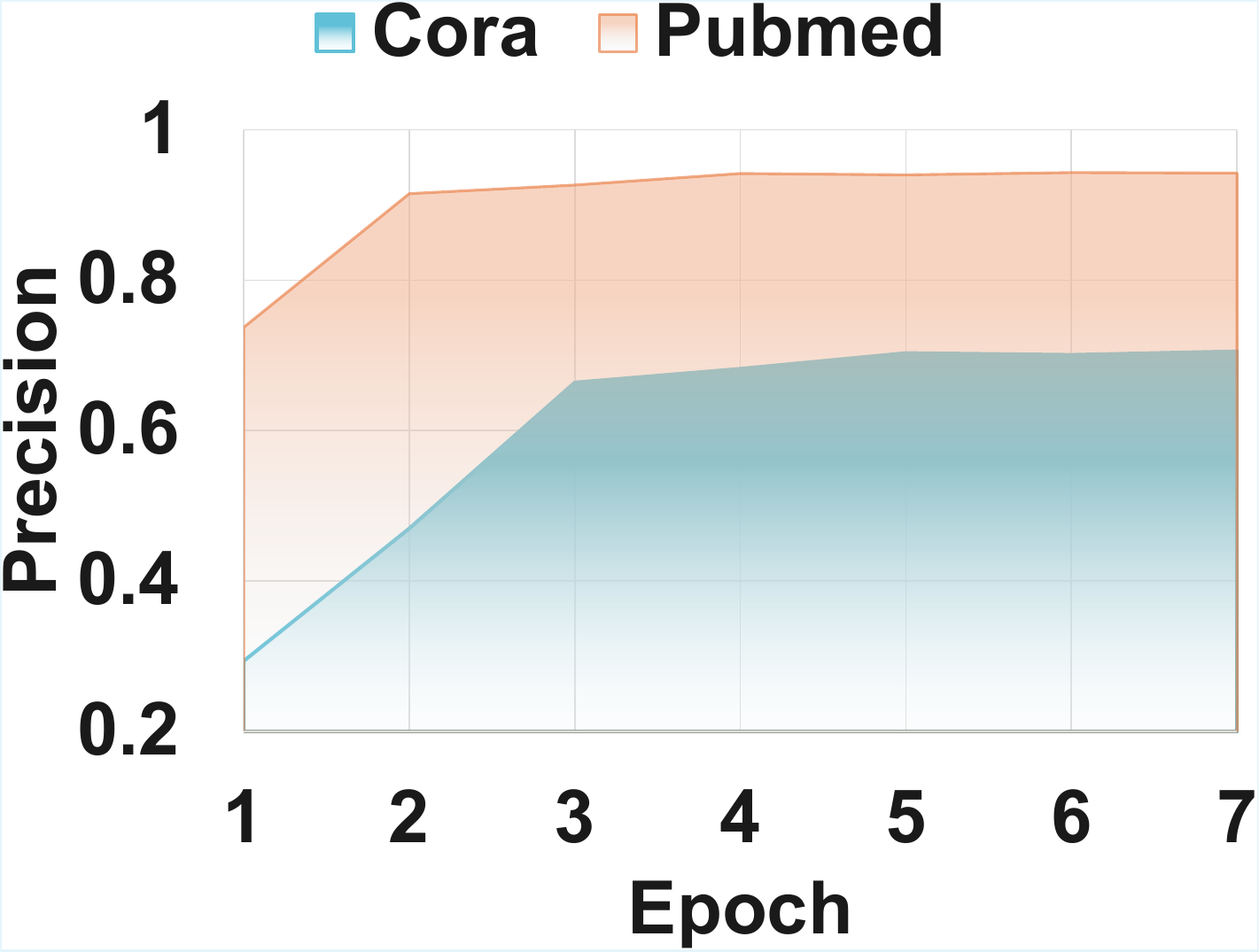}
        \caption{Precision of filtered samples over training epochs.}
        \label{fig:filter}
    \end{minipage}
\end{figure*}

\noindent \textbf{Evaluation Protocols and Metrics.} We adopt two evaluation protocols. Multimodal models such as \model{} are evaluated via zero-shot node classification, using cosine similarity between node and class embeddings. For graph SSL baselines, we follow a clustering-based protocol: applying K-Means on frozen embeddings and aligning clusters with ground-truth labels via the Hungarian algorithm. This avoids additional supervised training and measures the intrinsic embedding quality. We report ACC and F1-score for classification, AUC for link prediction, and MRR/Recall@K for retrieval.

\noindent \textbf{Ablation Studies.} To examine the contribution of each component, we construct three simplified variants of \model: \textbf{\model-CLIP:} trained only with the standard CLIP loss; \textbf{w/o Assessment:} disables the quality-aware assessment mechanism; \textbf{w/o Filter:} removes dynamic sample filtering while retaining loss modulation. These variants isolate the effects of dynamic adaptation and filtering.

\subsection{Unsupervised Graph–Text Alignment (RQ1)}
\subsubsection{Node Classification}
\label{subsec:unsupervised_node_class} 
In this setting, \model is composed of a 2-layer GCN as the graph encoder and a Qwen3-0.6B language model as the text encoder, trained from scratch.

\model{} demonstrates consistently strong performance across the six TAGs, as shown in Table~\ref{tab:unsupervised_acc_combined}. On clean data, ADAligner significantly surpasses nearly all unimodal baselines and existing multimodal approaches such as G2P2 and ConGraT. It also clearly improves upon the \model-Base variant, which relies solely on the standard CLIP loss, confirming that our many-to-many objectives capture richer semantic relationships and enhance expressive power. The advantages of \model{} become even more pronounced under noisy conditions: while other multimodal methods degrade substantially due to their static alignment strategies, \model{} maintains strong robustness and continues to outperform them by a large margin. These results validate the effectiveness of our dynamic quality-aware mechanism in mitigating noise and preserving high-quality representations, further supported by our ablation studies. We additionally report F1-scores in Appendix~\ref{app:Unsupervised Setting}, which highlight \model's balanced performance across clean and noisy settings.

\begin{table*}[t]
\centering
\captionsetup{skip=1pt}
\caption{Zero-shot node classification ACC for transferability assessment pre-trained on clean datasets and noisy datasets (30\% noise). The highest results are highlighted in \textbf{bold} and the second-best results are \underline{underlined}. F1-score is reported in Table~\ref{tab:app_zeroshot_f1_combined}.}
\label{tab:transferability_acc}
\resizebox{\textwidth}{!}{%
\begin{tabular}{lcccccccccccccccc}
\toprule
\multirow{2}{*}{Methods} 
& \multicolumn{2}{c}{Cora} 
& \multicolumn{2}{c}{Citeseer} 
& \multicolumn{2}{c}{WikiCS} 
& \multicolumn{2}{c}{Instagram} 
& \multicolumn{2}{c}{Ele-Photo} 
& \multicolumn{2}{c}{Ele-Computers} 
& \multicolumn{2}{c}{Ele-History} \\
\cmidrule(lr){2-3} \cmidrule(lr){4-5} \cmidrule(lr){6-7} \cmidrule(lr){8-9} 
\cmidrule(lr){10-11} \cmidrule(lr){12-13} \cmidrule(lr){14-15}
& 0\% & 30\% & 0\% & 30\% & 0\% & 30\% & 0\% & 30\% & 0\% & 30\% & 0\% & 30\% & 0\% & 30\% \\
\midrule
DGI\cite{dgi} & $14.24_{\pm4.77}$ & -- & $16.05_{\pm4.17}$ & -- & $6.10_{\pm4.09}$ & -- & $59.73_{\pm1.31}$ & -- & $10.98_{\pm3.83}$ & -- & $14.94_{\pm2.46}$ & -- & $7.58_{\pm2.93}$ & -- \\
GRACE\cite{grace} & $15.25_{\pm3.02}$ & -- & $14.98_{\pm4.76}$ & -- & $10.98_{\pm2.08}$ & -- & $52.08_{\pm3.12}$ & -- & $9.42_{\pm2.20}$ & -- & $9.74_{\pm5.39}$ & -- & $18.06_{\pm2.90}$ & -- \\
BGRL\cite{bgrl} & $16.06_{\pm2.01}$ & -- & $18.71_{\pm1.58}$ & -- & $13.26_{\pm3.05}$ & -- & $58.53_{\pm1.59}$ & -- & $8.65_{\pm2.99}$ & -- & $11.53_{\pm3.48}$ & -- & $9.45_{\pm3.83}$ & -- \\
GraphMAE\cite{graphmae} & $18.78_{\pm1.88}$ & -- & $20.69_{\pm3.12}$ & -- & $9.08_{\pm1.40}$ & -- & $60.49_{\pm1.87}$ & -- & $5.01_{\pm1.80}$ & -- & $13.54_{\pm3.94}$ & -- & $18.66_{\pm3.94}$ & -- \\
ZeroG\cite{li2024zerog} & $42.63_{\pm2.15}$ & $36.97_{\pm1.16}$ & $29.71_{\pm2.69}$ & $27.43_{\pm1.87}$ & $17.48_{\pm1.19}$ & $18.48_{\pm2.01}$ & $55.63_{\pm1.38}$ & $53.63_{\pm2.15}$ & $32.19_{\pm3.85}$ & $30.12_{\pm1.32}$ & $40.31_{\pm2.48}$ & $30.20_{\pm3.69}$ & $30.68_{\pm2.65}$ & $28.27_{\pm1.94}$ \\
\midrule
G2P2\cite{wen2023augmenting} & $28.30_{\pm2.93}$ & $11.18_{\pm4.83}$ & $23.84_{\pm3.75}$ & $17.21_{\pm2.01}$ & $17.36_{\pm8.42}$ & $13.47_{\pm7.88}$ & $53.51_{\pm7.64}$ & $60.05_{\pm3.19}$ & $9.56_{\pm6.43}$ & $13.68_{\pm8.87}$ & $18.36_{\pm7.01}$ & $11.95_{\pm5.18}$ & $10.40_{\pm3.74}$ & $6.08_{\pm4.21}$ \\
GraphCLIP\cite{zhu2025graphclip} & $52.79_{\pm1.58}$ & $51.02_{\pm4.06}$ & $48.92_{\pm3.39}$ & $48.13_{\pm4.29}$ & $31.14_{\pm1.63}$ & $27.25_{\pm1.92}$ & $\mathbf{62.96}_{\pm0.18}$ & $\mathbf{63.18}_{\pm0.17}$ & $40.03_{\pm0.89}$ & $41.63_{\pm0.26}$ & $40.49_{\pm1.67}$ & $35.43_{\pm4.12}$ & $51.29_{\pm2.36}$ & $54.40_{\pm1.76}$ \\
\midrule
\rowcolor{mygray}
\model & $\mathbf{66.91}_{\pm2.20}$ & $\mathbf{66.17}_{\pm3.17}$ & $\mathbf{62.75}_{\pm3.30}$ & $\mathbf{60.94}_{\pm1.95}$ & $\mathbf{49.17}_{\pm4.19}$ & $\mathbf{52.96}_{\pm2.96}$ & $62.22_{\pm1.61}$ & $\underline{62.70}_{\pm0.78}$ & $\mathbf{51.81}_{\pm1.42}$ & $\mathbf{50.41}_{\pm1.77}$ & $\mathbf{54.07}_{\pm2.26}$ & $\mathbf{53.02}_{\pm2.46}$ & $\mathbf{58.13}_{\pm1.35}$ & $\mathbf{58.72}_{\pm1.63}$ \\
\model-CLIP & $65.58_{\pm2.34}$ & $64.68_{\pm1.35}$ & $59.50_{\pm0.68}$ & $59.97_{\pm0.80}$ & $44.32_{\pm3.27}$ & $42.71_{\pm4.11}$ & $61.66_{\pm0.15}$ & $61.54_{\pm1.63}$ & $39.49_{\pm1.57}$ & $44.79_{\pm0.71}$ & $48.12_{\pm2.09}$ & $51.63_{\pm1.46}$ & $54.60_{\pm1.02}$ & $56.14_{\pm0.94}$ \\
w/o Assessment & $65.22_{\pm1.32}$ & ${64.76}_{\pm1.73}$ & ${61.04}_{\pm1.79}$ & ${60.23}_{\pm0.95}$ & ${45.81}_{\pm1.52}$ & $50.26_{\pm4.91}$ & $\underline{62.59}_{\pm0.28}$ & $61.69_{\pm1.03}$ & $44.57_{\pm0.58}$ & ${48.40}_{\pm0.57}$ & $49.91_{\pm1.81}$ & $\underline{52.70}_{\pm1.22}$ & $56.18_{\pm1.18}$ & $57.48_{\pm1.23}$ \\
w/o Filter & $\underline{65.62}_{\pm1.99}$ & $\underline{65.13}_{\pm1.79}$ & $\underline{61.89}_{\pm1.66}$ & $\underline{60.85}_{\pm2.11}$ & $\underline{48.96}_{\pm2.04}$ & $\underline{51.37}_{\pm2.46}$ & $61.13_{\pm1.87}$ & $62.56_{\pm0.95}$ & $\underline{51.14}_{\pm1.59}$ & $\underline{49.52}_{\pm1.21}$ & $\underline{53.70}_{\pm1.63}$ & $52.12_{\pm2.41}$ & $\underline{57.68}_{\pm1.74}$ & $\underline{58.22}_{\pm2.58}$ \\
\bottomrule
\end{tabular}}
\end{table*}

\begin{table}[!htbp]
\centering
\captionsetup{skip=2pt}
\begin{minipage}[b]{0.49\columnwidth} 
    \centering
    \captionof{table}{Robustness to encoder choices on WikiCS.}
    \label{tab:wikics_perf}
    \resizebox{\linewidth}{!}{ 
    \begin{tabular}{@{}lcc@{}}
        \toprule
        Methods & 0\%  & 30\% \\ 
        \midrule
        Variant 1-CLIP & $56.19_{\pm2.82}$ & $53.67_{\pm2.60}$ \\
        Variant 1 & $\mathbf{63.05}_{\pm1.87}$ & $\mathbf{61.81}_{\pm2.90}$ \\ \hline
        Variant 2-CLIP & $59.96_{\pm0.44}$ & $56.60_{\pm1.79}$ \\
        Variant 2 & $\mathbf{61.21}_{\pm2.18}$ & $\mathbf{60.69}_{\pm0.74}$ \\
        \bottomrule
    \end{tabular}
    }
\end{minipage}
\hfill
\begin{minipage}[b]{0.49\columnwidth} 
    \centering
    \captionof{table}{Pre-training efficiency v.s. average ACC.}
    \label{tab:training_time_comparison_gfm}
    \resizebox{\linewidth}{!}{ 
    \begin{tabular}{@{}lcc@{}}
        \toprule
        Methods & Time (s) & Avg. ACC (\%) \\
        \midrule
        G2P2 & 18014.16 & 30.02 \\
        GraphCLIP & 12619.41 & 48.80 \\
        \rowcolor{mygray}
        \model & $\mathbf{5197.42}$ & $\mathbf{58.32}$ \\
        \bottomrule
    \end{tabular}
    }
\end{minipage}
\end{table}

To further verify that the improvements of \model{} are not tied to specific encoder architectures, we replace the default encoders with alternative graph and text backbones (\textbf{Variant 1}: GraphSAGE + Qwen3-0.6B, \textbf{Variant 2}: GCN + SBERT) and re-train under the same setting on WikiCS. As shown in Table~\ref{tab:wikics_perf}, the proposed dynamic mechanism consistently yields clear gains over the corresponding CLIP-only baselines across all encoder combinations, under both clean and noisy conditions. This demonstrates that \model{} generalizes robustly across backbones, confirming the broad applicability of its adaptive alignment strategy beyond a particular architecture.

\subsection{Transfer to Unseen Graphs/Tasks (RQ2)}
To isolate and evaluate our dynamic loss in the transfer setting, we simply replace the original alignment loss within the GraphCLIP framework, using its encoders as a fixed backbone and maintaining its original hyperparameter settings for a controlled comparison.

\subsubsection{Zero-Shot Node Classification}
\label{subsec:transferability}
\model{} markedly enhances generalization and robustness. As shown in Table~\ref{tab:transferability_acc}, \model{} achieves superior ACC on nearly all target datasets. Notably, even when GraphCLIP attains slightly higher ACC on Instagram, \model's F1 score is substantially better, indicating more balanced predictions. Under noisy pre-training, this advantage becomes more pronounced, with \model{} maintaining strong ACC and F1 while other multimodal baselines degrade significantly. These confirm our dynamic loss learns more robust and transferable representations. We further complement these results with ablation analysis (Appendix~\ref{app:app_zero_shot}), which shows both dynamic assessment and filter are indispensable.

Beyond accuracy, \model{} also offers improved efficiency. As shown in Table~\ref{tab:training_time_comparison_gfm}, its pre-training requires less than half the time of GraphCLIP while yielding nearly a 10-point higher average ACC. This efficiency stems from fundamentally different robustness strategies: GraphCLIP relies on costly adversarial perturbations, whereas \model{} achieves robustness more efficiently by dynamically adapting its objective and filtering noisy data. Thus, \model{} not only enhances robustness and transferability, but also achieves these improvements with substantially higher efficiency.

\subsubsection{Few-Shot Prompting}
\label{subsec:few_shot}
To further assess transferability under limited supervision, we evaluate \model{} with a standard few-shot prompting setup (1, 3, 5, and 10-shot). As shown in Figure~\ref{fig:few_shot_results_wikics}, when pre-trained on clean data, \model{} achieves results comparable to GraphCLIP, showing no substantial gap between the two. However, under the more challenging noisy pre-training condition, \model{} consistently and clearly outperforms GraphCLIP across all shot settings. This contrast underscores the key advantage of our dynamic loss: while maintaining competitiveness under clean supervision, it delivers substantially stronger robustness when noise is present, which is crucial for realistic few-shot applications.

\begin{figure*}[!t] 
\centering 
\captionsetup{skip=1pt}

\begin{subfigure}[b]{0.24\textwidth}
    \centering
    \includegraphics[width=\textwidth]{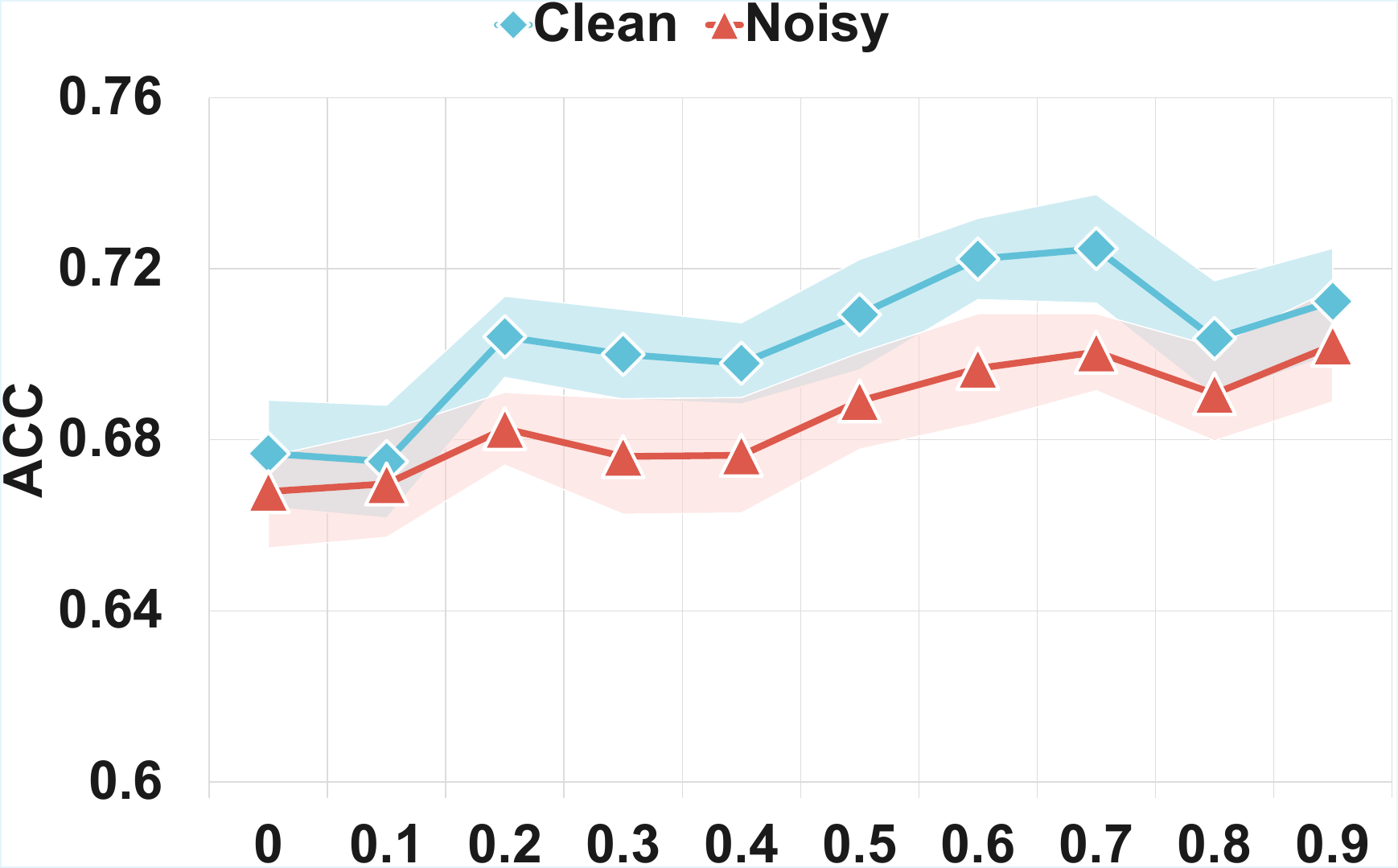}
    \caption{Parameter $\alpha$}
    \label{fig:sub_alpha}
\end{subfigure}
\hfill 
\begin{subfigure}[b]{0.24\textwidth}
    \centering
    \includegraphics[width=\textwidth]{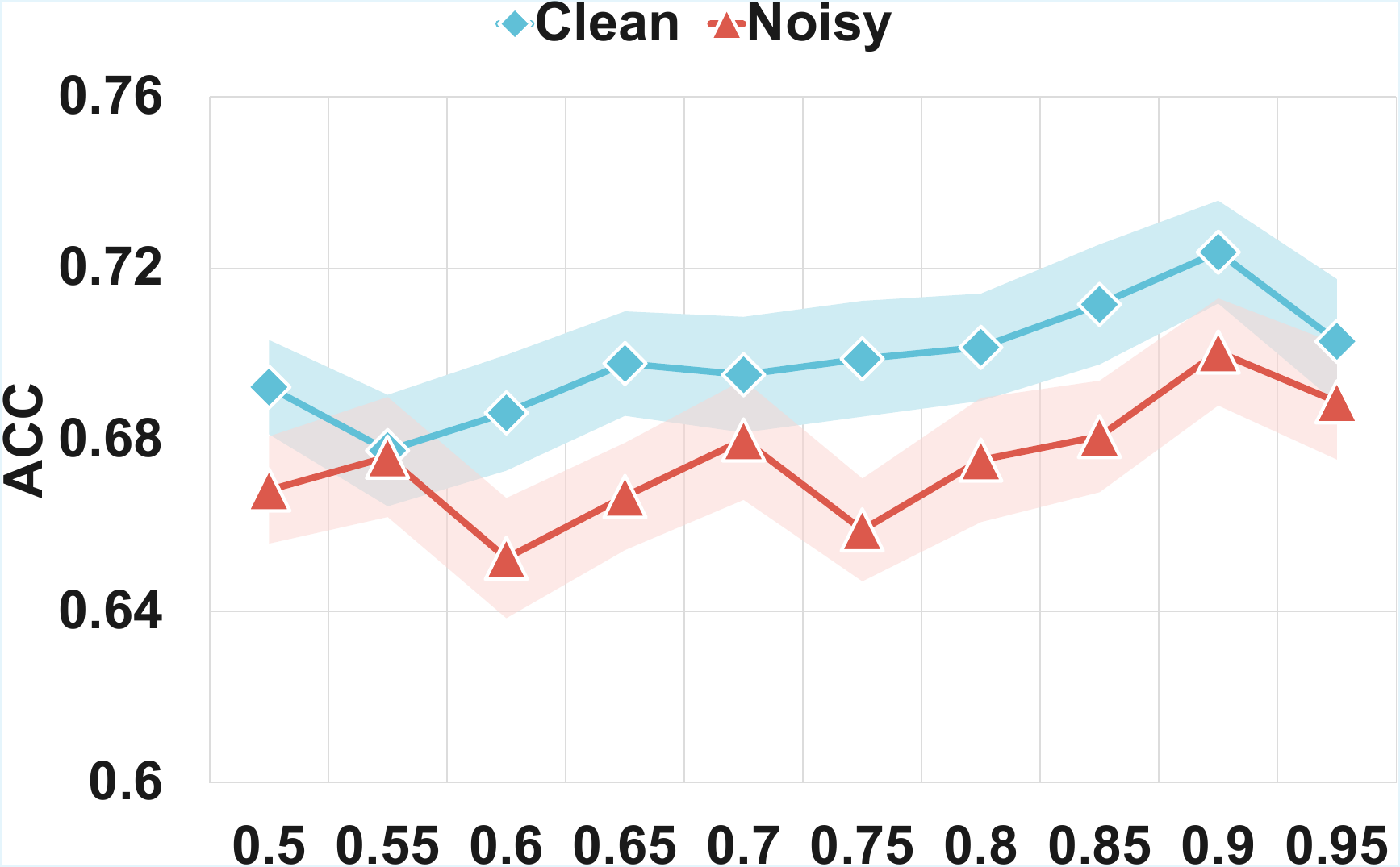}
    \caption{Parameter $m$}
    \label{fig:sub_m}
\end{subfigure}
\hfill
\begin{subfigure}[b]{0.24\textwidth}
    \centering
    \includegraphics[width=\textwidth]{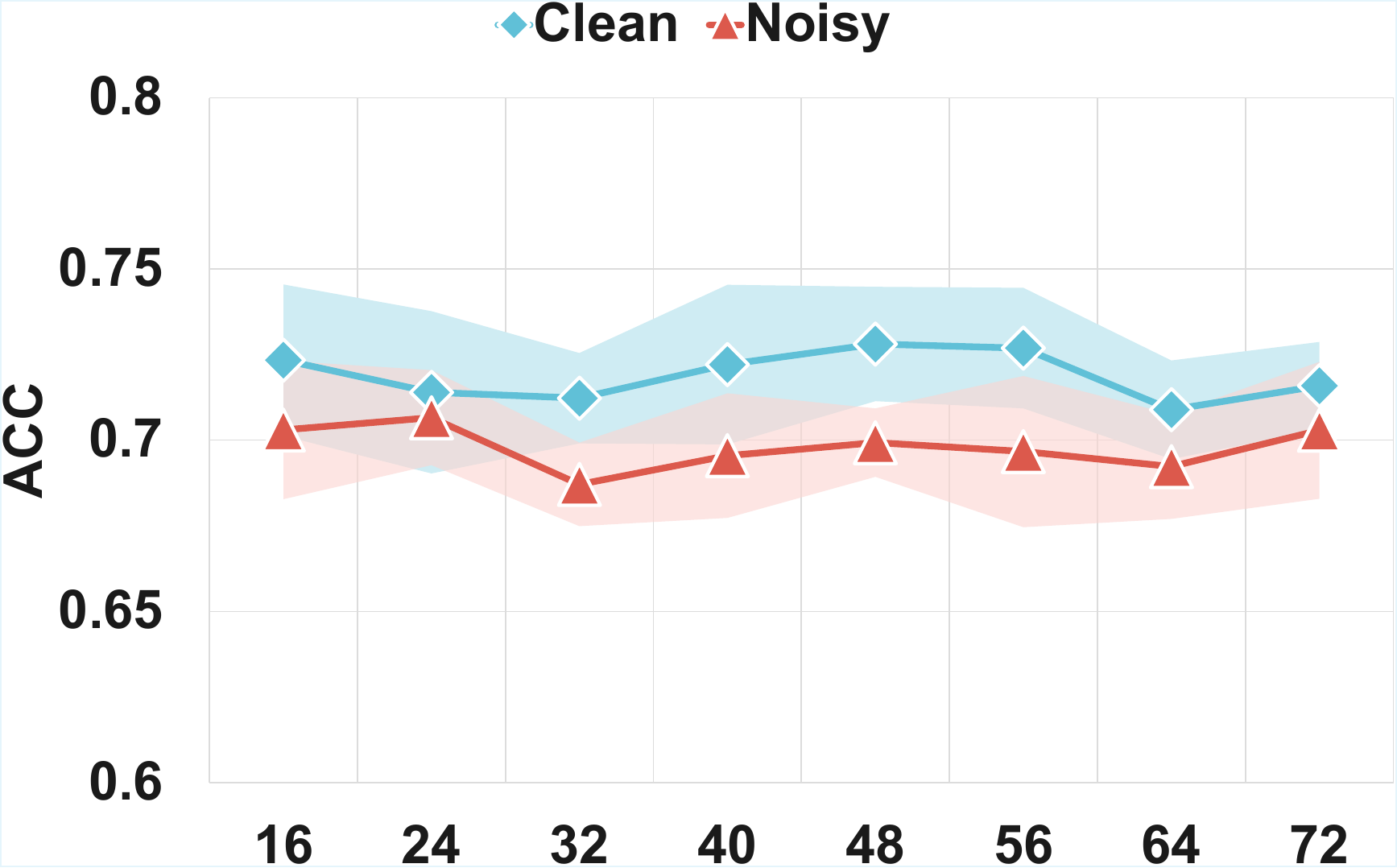}
    \caption{Batch size on Cora}
    \label{fig:sub_bs_cora}
\end{subfigure}
\hfill
\begin{subfigure}[b]{0.24\textwidth}
    \centering
    \includegraphics[width=\textwidth]{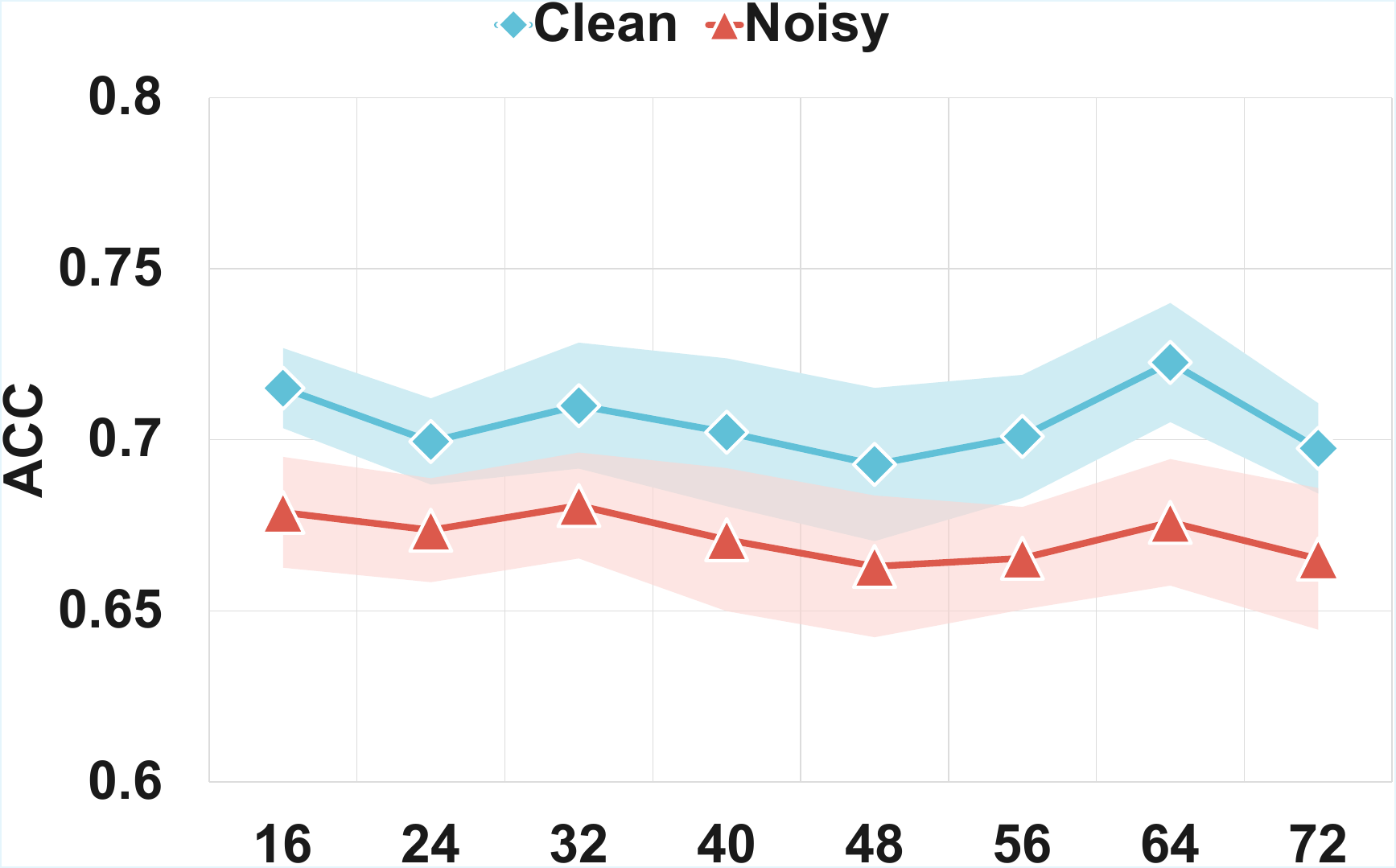}
    \caption{Batch size on CiteSeer}
    \label{fig:sub_bs_citeseer}
\end{subfigure}

\caption{Sensitivity analysis of hyperparameters for \model under clean and noisy (30\%) settings.}
\label{fig:sensitivity_analysis_all} 
\end{figure*}

\subsubsection{Link Prediction}
\label{subsec:link_prediction}
We also evaluate link prediction to test whether \model{} preserves structural information in its embeddings. 
As shown in Table~\ref{tab:link_prediction_clean_noisy}, \model{} generally achieves higher AUC scores than GraphCLIP across most datasets. In particular, the gains are substantial on Cora and WikiCS, demonstrating that our dynamic loss better captures structural cues that support link reasoning. On History, ADAligner outperforms GraphCLIP on clean data and remains competitive under noise. Overall, these findings indicate that \model{} not only enhances semantic alignment but also maintains strong structural representations.

\begin{table}[h] 
  \centering 
  \captionsetup{skip=1pt}
  \caption{Link prediction results under clean and noisy (30\%).}
  \label{tab:link_prediction_clean_noisy} 
  \resizebox{1.0\linewidth}{!}{
  \scriptsize

  \setlength{\tabcolsep}{4pt} 

  \begin{tabular}{@{}l*{6}{c}@{}}
    \toprule
    \multirow{2}{*}{Methods} & \multicolumn{2}{c}{Cora} & \multicolumn{2}{c}{WikiCS} & \multicolumn{2}{c}{History} \\
    \cmidrule(lr){2-3} \cmidrule(lr){4-5} \cmidrule(lr){6-7}
    & 0\% & 30\% & 0\% & 30\% & 0\% & 30\% \\
    \midrule
    GraphCLIP   & $73.85_{\pm 1.71}$ & $80.67_{\pm 2.93}$ & $72.42_{\pm 1.44}$ & $74.78_{\pm 2.66}$ & $88.29_{\pm 0.94}$ & $\bm{91.04}_{\pm 0.84}$ \\
    \rowcolor{mygray}
    \model & $\bm{84.99}_{\pm 1.57}$ & $\bm{85.06}_{\pm 0.97}$ & $\bm{82.81}_{\pm 1.92}$ & $\bm{82.48}_{\pm 0.68}$ & $\bm{90.34}_{\pm 1.20}$ & $90.53_{\pm 1.39}$ \\
    \bottomrule
  \end{tabular}
  }
\end{table}

\subsubsection{Retrieval Task}
\label{subsec:retrieval} 
We assess cross-modal retrieval performance on the WikiCS dataset, encompassing both node-to-text (N2T) and text-to-node (T2N) tasks. Performance is evaluated under two scopes: global retrieval and category-specific retrieval.

\model substantially improves cross-modal retrieval, particularly for N2T tasks on WikiCS, as shown in Table~\ref{tab:retrieval_results_clean_pretrained}. Regardless of whether pre-trained on clean or 30\% noisy source data, \model consistently yields significant N2T performance gains over the original GraphCLIP. Strong, competitive results are also observed for T2N retrieval. These findings underscore that \model's dynamic loss cultivates a more effective and robust graph-text alignment. Full results are detailed in Appendix~\ref{app:app_retrieval}.

\begin{table}[!htbp]
\centering
\captionsetup{skip=1pt}
\caption{Global retrieval results on clean data.}
\resizebox{0.9\linewidth}{!}{
\begin{tabular}{@{}llcccc@{}}
    \toprule
    Methods & Task & MRR & R@1 & R@5 & R@10 \\
    \midrule
    \multirow{2}{*}{GraphCLIP} 
        & N2T & $5.95 \pm 1.60$ & $3.64 \pm 1.05$ & $7.85 \pm 2.10$ & $10.49 \pm 2.68$ \\
        & T2N & $86.47 \pm 5.17$ & $80.42 \pm 6.48$ & $93.97 \pm 3.65$ & $96.45 \pm 2.73$ \\
    \midrule
    \multirow{2}{*}{\textbf{\model}} 
        & N2T & \textbf{36.68} $\pm$ \textbf{3.50} & \textbf{28.02} $\pm$ \textbf{2.95} & \textbf{46.00} $\pm$ \textbf{4.29} & \textbf{54.11} $\pm$ \textbf{4.74} \\
        & T2N & \textbf{90.40} $\pm$ \textbf{2.89} & \textbf{85.59} $\pm$ \textbf{4.15} & \textbf{96.46} $\pm$ \textbf{1.33} & \textbf{98.08} $\pm$ \textbf{0.83} \\
    \bottomrule
\end{tabular}
}
\label{tab:retrieval_results_clean_pretrained}
\end{table}

\subsection{Analysis of the Dynamic Mechanism (RQ3)}
\subsubsection{Dynamic Controller}
\textbf{Our fixed-$\theta$ analysis confirms that the controller adapts the learning objective to data quality, thereby ensuring robustness.} To validate the controllability of the dynamic mechanism, we conduct a fixed-$\theta$ analysis on the Cora dataset, setting $\theta$ to $\{0.9, 1.0, 1.1\}$ and evaluating zero-shot node classification ACC under clean and noisy conditions.

The results in Figure~\ref{fig:fixed_theta_performance_en} show a clear trend: on clean data, a higher fixed $\theta$ yields better performance, consistent with emphasizing expressive many-to-many objectives; under noisy data, a lower fixed $\theta$ performs better, reflecting the need to prioritize robust one-to-one alignment. This contrasting behavior provides direct empirical evidence for the intended negative feedback loop of our controller, further supporting the stability guarantees established in Theorem~\ref{thm:convergence_main}.

\subsubsection{Dynamic Filter}
\label{subsec:filter}
\textbf{The dynamic filter enhances robustness by progressively learning to identify and discard genuine noisy pairs during training.} 
As training progresses, the filter is expected to remove truly mismatched graph–text pairs rather than mistakenly discarding hard positives. 
For transparency, we measure precision by directly comparing filtered-out pairs with ground-truth synthetic noise labels. 
Figure~\ref{fig:filter} shows that the precision steadily increases across epochs on Cora and Pubmed, reaching a high level in later stages. 
This indicates that the filter becomes increasingly selective and adaptively targets genuine noisy pairs, thereby playing a critical role in maintaining \model’s robustness under noisy conditions.

\subsection{Hyperparameter Sensitivity Analysis (RQ4)}
\label{subsec:sensitivity_analysis}

\textbf{\model remains stable across a broad range of its key hyperparameters.} We first analyze sensitivity of two factors: the quality deviation sensitivity parameter $\alpha$ and the EMA momentum parameter $m$. Each parameter is varied across a wide range while keeping others fixed at their default values ($\alpha=0.7$, $m=0.9$). Experiments are conducted on Cora under both clean and noisy conditions, using zero-shot node classification accuracy for evaluation.  

Figure~\ref{fig:sub_alpha} and Figure~\ref{fig:sub_m} show that \model{} maintains reliable performance across a broad spectrum of hyperparameter values. For $\alpha$, extremely small values hinder adaptation and degrade accuracy, while excessively large values risk overreacting to noise; the default $\alpha=0.7$ provides a balanced trade-off. For $m$, lower values lead to instability due to overly reactive historical statistics, whereas higher values yield smoother and more robust performance, with $m=0.9$ striking a good balance between stability and responsiveness.

Our analysis of batch size sensitivity on Cora and CiteSeer (Figures~\ref{fig:sub_bs_cora}–\ref{fig:sub_bs_citeseer}) further confirms the model's robustness. \model{} maintains stable accuracy across a wide range of batch sizes with only minor fluctuations, even under noisy conditions. This stability highlights the dynamic mechanism's effectiveness in adapting to varying batch statistics, confirming that reliable performance is achievable without extensive hyperparameter tuning.

\section{Related Work}

\noindent \textbf{Graph--Text Alignment.}
Contrastive learning frameworks, inspired by CLIP~\cite{clip}, have become the dominant paradigm for aligning graph structures with textual attributes on TAGs. Methods like GraphCLIP~\cite{zhu2025graphclip}, G2P2~\cite{wen2023augmenting}, and GRENADE~\cite{li2023grenade} adopt a dual-encoder design, but their alignment strategies rely on strict one-to-one positive pairs, making it difficult to capture the many-to-many relations commonly observed in real-world graphs. To alleviate this, follow-up works—drawing on ideas from vision–language models like SoftCLIP~\cite{gao2024softclip}—introduced soft-alignment objectives to model multiple correspondences. For example, ConGraT~\cite{brannon2024congrat} incorporates graph similarity scores into a fixed many-to-many alignment target. However, these approaches are static and cannot adapt to varying data quality during training. Another line of research explores integrating pre-trained language models more directly, such as GLEM~\cite{glem}, which combines LLMs and GNNs for joint representation learning. In contrast to these static or hybrid strategies, \model{} adaptively balances one-to-one and many-to-many alignment objectives in real time, guided by batch-level quality estimation.  

\noindent \textbf{Adaptive and Self-Training in Representation Learning.}
Robust representation learning often employs adaptation to data quality or model states. A classical paradigm is self-training, where models use high-confidence predictions as pseudo-labels while filtering out uncertain samples to avoid error propagation~\cite{li2022informative,wang2024distribution,sun2020multistageselfsupervisedlearninggraph}. Related ideas appear in curriculum learning~\cite{bengio2009curriculum} and MentorNet~\cite{jiang2018mentornet}, where sample selection or weighting evolves dynamically over training. These principles of confidence-based filtering and dynamic weighting have been extended to various robust learning setups, yet prior methods typically treat sample selection and loss reweighting as separate processes. By contrast, \model{} unifies them under a single adaptive controller: a batch-level quality score governs both which samples are retained and how one-to-one versus many-to-many losses are combined. This integration enables \model{} to adjust supervision signals on the fly, promoting semantic diversity when data is clean while maintaining robustness when it is noisy.  

\section{Conclusion}
\label{sec:conclusion}
In this work, we addressed the core trade-off in graph–text alignment: balancing expressive but noise-vulnerable many-to-many objectives against robust but restrictive one-to-one strategies. To overcome this challenge, we proposed \model, a dynamic quality-aware framework that performs real-time batch quality assessment and adaptively modulates both its alignment objectives and sample filtering strategy. This design enables the model to emphasize expressive many-to-many alignment on clean data, while pivoting to conservative one-to-one objectives with stricter filtering under noisy supervision. Extensive experiments across diverse TAG benchmarks demonstrate that \model achieves strong zero-/few-shot performance, robustness to alignment noise, and $2$--$3\times$ pre-training speedups over baselines, highlighting its effectiveness in both clean and noisy scenarios.


\bibliographystyle{ACM-Reference-Format}
\bibliography{references}

\appendix

\section{Theoretical Analysis}
\label{sec:theoretical_analysis}

We now provide the detailed proofs for the results stated in Section~\ref{subsec:theoretical_analysis}. 
We begin with the boundedness of key signals, then establish the stability of the controller, and finally analyze the convergence of the total loss.

\begin{figure*}[htbp]
    \centering
    \captionsetup{skip=3pt}

    \begin{subfigure}[b]{0.24\textwidth}
        \centering
        \includegraphics[width=\linewidth]{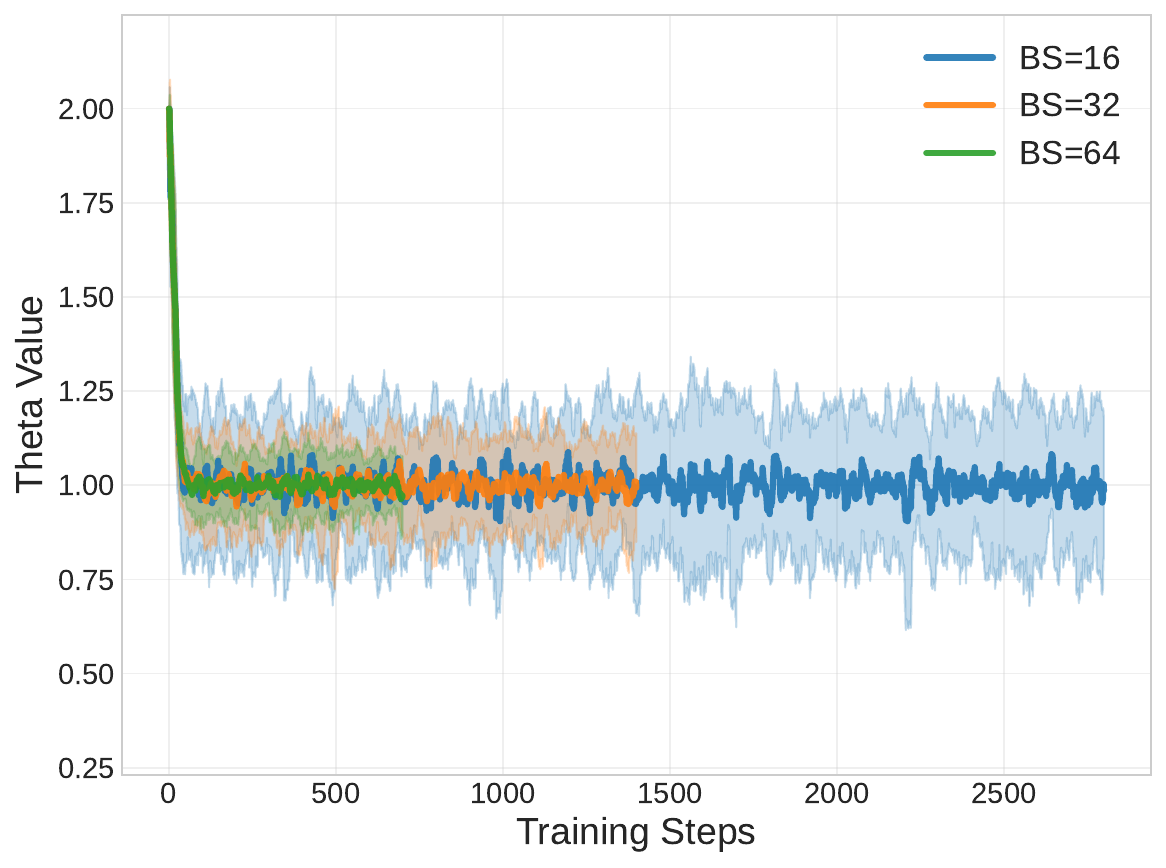}
        \caption{Batch Size (Clean)}
        \label{fig:citeseer_bs_clean}
    \end{subfigure}
    \hfill
    \begin{subfigure}[b]{0.24\textwidth}
        \centering
        \includegraphics[width=\linewidth]{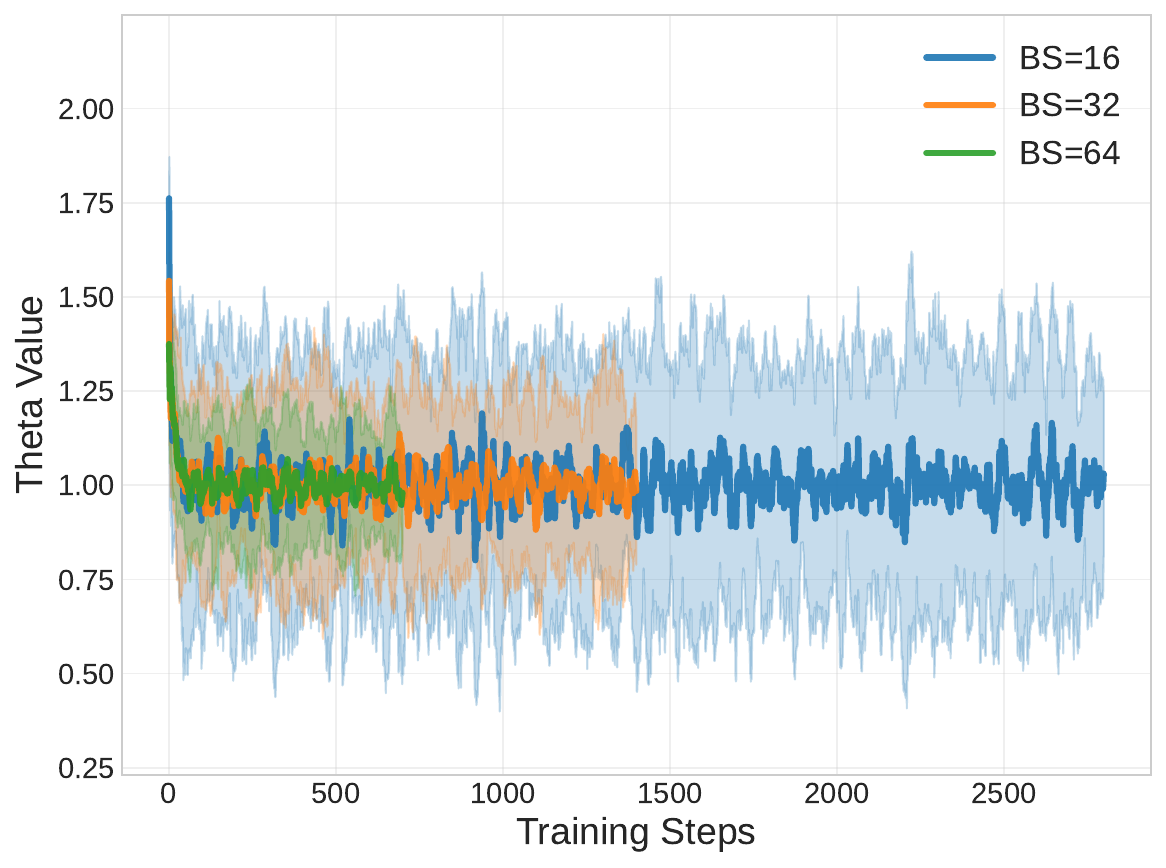}
        \caption{Batch Size (Noisy)}
        \label{fig:citeseer_bs_noisy}
    \end{subfigure}
    \hfill
    \begin{subfigure}[b]{0.24\textwidth}
        \centering
        \includegraphics[width=\linewidth]{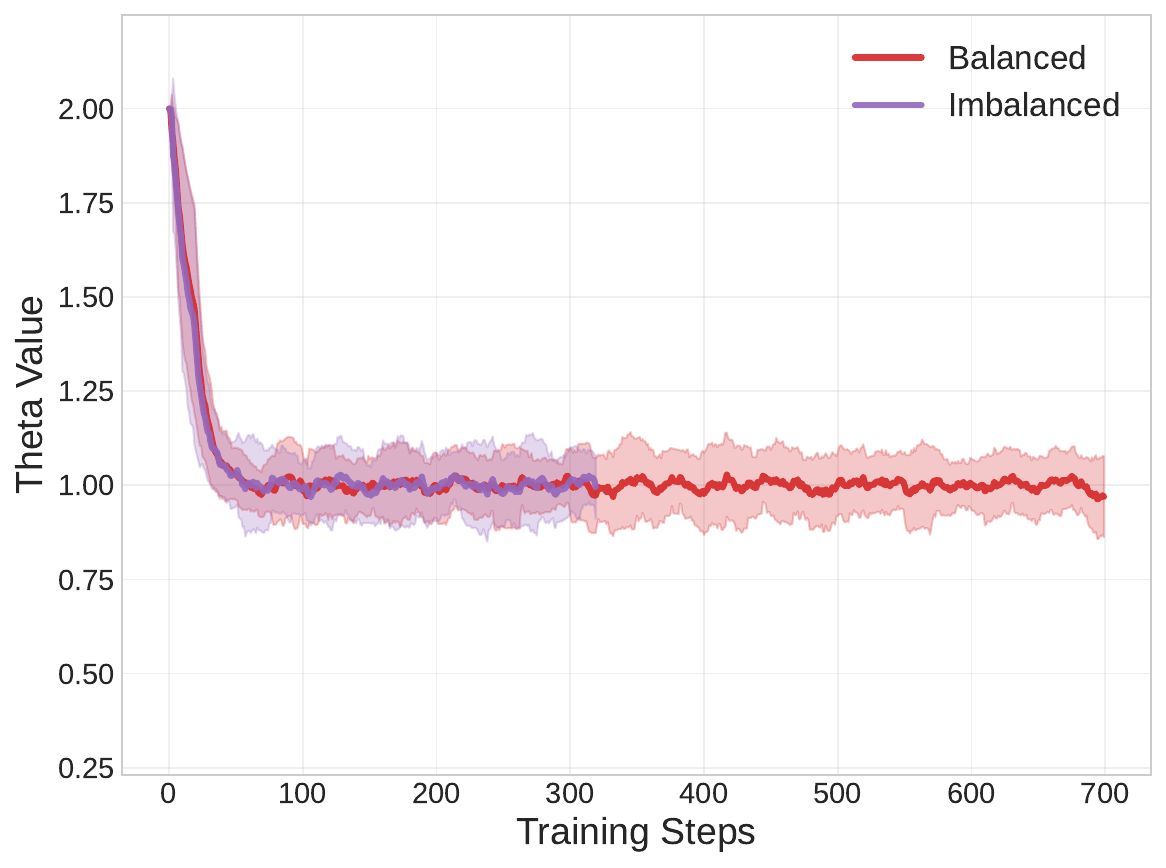}
        \caption{Class Imbalance (Clean)}
        \label{fig:citeseer_imbalance_clean}
    \end{subfigure}
    \hfill
    \begin{subfigure}[b]{0.24\textwidth}
        \centering
        \includegraphics[width=\linewidth]{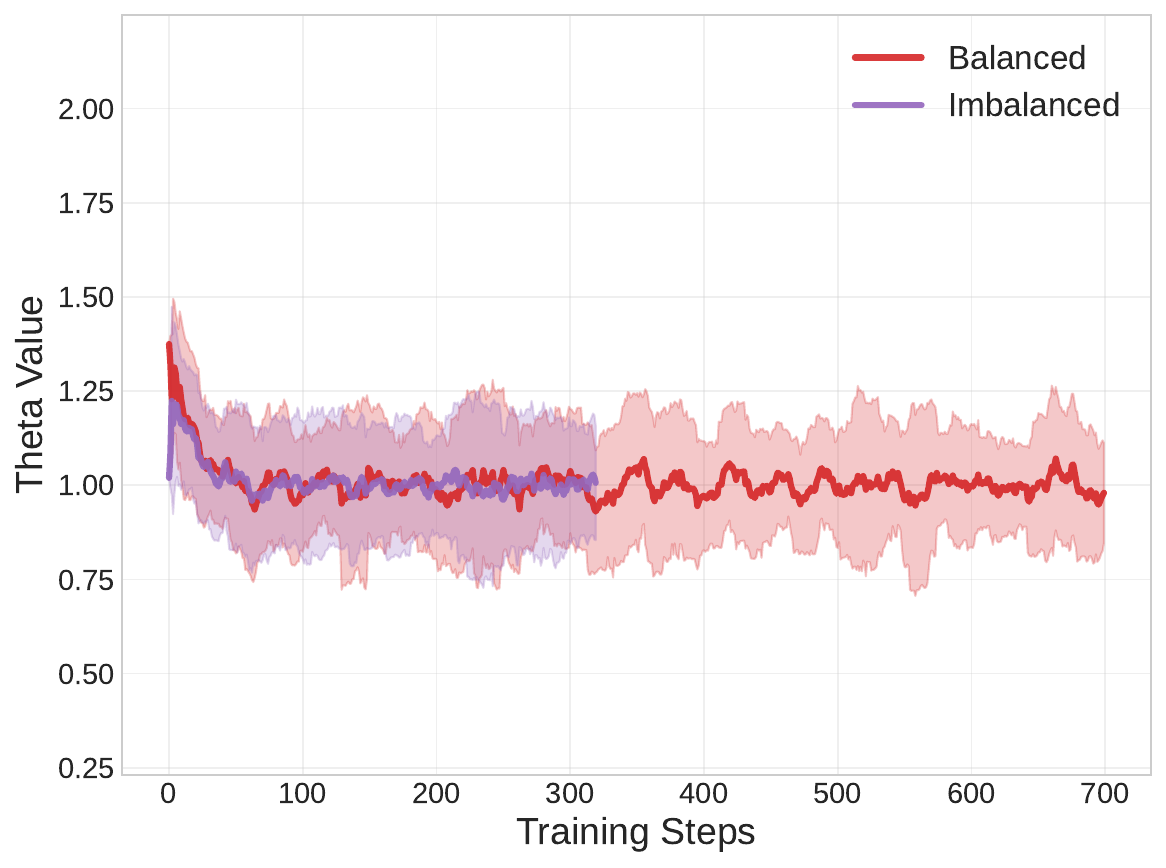}
        \caption{Class Imbalance (Noisy)}
        \label{fig:citeseer_imbalance_noisy}
    \end{subfigure}

    \caption{Empirical stability analysis of the control factor \(\theta\) on the \textbf{Citeseer} dataset. 
    The first two figures show the effect of varying batch sizes (\(BS \in \{16, 32, 64\}\)) on clean (0\% noise) and noisy (30\% noise) data. 
    The last two figures show the effect of class imbalance under clean and noisy conditions, with imbalance generated via long-tail sampling.}
    \label{fig:theta_stability_citeseer}
\end{figure*}

\subsection{Proof of Lemma~\ref{lem:boundedness}}
\begin{proof}
1. For any sample $i$ in batch $\mathcal B$, the quality score is
\[
M_i = \langle \hat{\mathbf g}_i,\hat{\mathbf t}_i\rangle 
- \tfrac{1}{N_B-1}\sum_{j\neq i}\langle \hat{\mathbf g}_i,\hat{\mathbf t}_j\rangle,
\]
where $\hat{\mathbf g}_i,\hat{\mathbf t}_j$ are L2-normalized. Since $\langle \cdot,\cdot\rangle\in[-1,1]$, we have $M_i\in[-2,2]$, hence $M_{\mathcal B}\in[-2,2]$.

2. The controller is updated as
\[
\theta = \theta_0 + \alpha(M_{\mathcal B}-M_0),
\]
where $M_0$ is an EMA of $M_{\mathcal B}$, also bounded in $[-2,2]$. Thus the unclamped $\theta$ lies in $[\theta_0-4\alpha,\,\theta_0+4\alpha]$. The implementation applies a projection $\Pi_{[\theta_{\min},\theta_{\max}]}$ with $\theta_{\min}>-1$, ensuring boundedness.

3. The dynamic weights are defined as
\[
\beta(\theta)=\theta\cdot\beta_0,\quad 
\gamma(\theta)=\theta\cdot\gamma_0,\quad
\mu(\theta)=\frac{\mu_0}{1+\theta}.
\]
Since $\theta$ is bounded and $\theta_{\min}>-1$, the denominator in $\mu(\theta)$ is bounded away from $0$. Therefore, all weights are bounded.
\end{proof}

\subsection{Proof of Lemma~\ref{lem:stability}}
\begin{proof}[Proof Sketch]
Let $\theta^*$ denote the equilibrium point. Define the Lyapunov function
\[
V_t = \mathbb E[(\theta_t-\theta^*)^2].
\]

The update of $\theta_t$ is
\[
\theta_{t+1} = \Pi\big(\theta_t+\alpha(M_{\mathcal B}(\Phi_t)-M_{0,t})\big).
\]

The EMA update is
\[
M_{0,t+1} = (1-\rho)M_{0,t} + \rho M_{\mathcal B}(\Phi_t).
\]

Using the first-order expansion
\[
M_{\mathcal B}(\Phi_{t+1}) \approx M_{\mathcal B}(\Phi_t) 
- \eta \langle \nabla_\Phi M_{\mathcal B}(\Phi_t),\,\nabla_\Phi \mathcal L(\Phi_t,\theta_t)\rangle,
\]
and the negative-drift condition
\[
\mathbb E[\langle \nabla_\Phi M_{\mathcal B}(\Phi_t),\,\nabla_\Phi \mathcal L(\Phi_t,\theta_t)\rangle]
\propto -(\theta_t-\theta^*),
\]
we obtain a recursion
\[
\mathbb E[(\theta_{t+1}-\theta^*)^2]
\le (1-c)\,\mathbb E[(\theta_t-\theta^*)^2] + C_1\eta^2 + C_2\alpha^2,
\]
with $c\propto \alpha\rho\kappa > 0$. 

Thus $\theta_t$ converges in expectation to an $\mathcal O(\eta+\alpha)$-neighborhood of $\theta^*$. The projection operator $\Pi$ is non-expansive, so stability is preserved.
\end{proof}

\subsection{Empirical Validation of Controller Stability}
\label{app:theta_stability}

To complement the theoretical analysis, we empirically evaluate the 
evolution of the controller $\theta$ across training steps under 
different experimental settings. Specifically, we vary the batch size 
($BS \in \{16,32,64\}$), introduce long-tail class imbalance, and 
compare clean versus noisy (30\% noise) data conditions. 

As shown in Figure~\ref{fig:theta_stability_citeseer}, $\theta$ consistently 
converges to a bounded neighborhood across all scenarios. This confirms 
our theoretical prediction of stability and demonstrates that the 
controller remains well-behaved even under challenging data 
distributions.

\subsection{Proof of Theorem~\ref{thm:convergence_main}}
\begin{proof}
The expected total loss is
\[
\begin{aligned}
\mathbb E[\mathcal L_{\text{total}}(\Phi_t,\theta_t)] 
= \mathbb E \Big[ & \beta(\theta_t)\mathcal L_{\text{soft}}(\mathcal B^f) 
+ \gamma(\theta_t)\mathcal L_{\text{CLIP}}^{\text{sub}}(\mathcal B^f) \\
& + \mu(\theta_t)\mathcal L_{\text{CLIP}}(\mathcal B^f)\Big].
\end{aligned}
\]

By the Descent Lemma for $L_\Phi$-smooth functions under unbiased gradient with variance $\sigma^2$, we have
\[
\begin{aligned}
\mathbb E[\mathcal L_{\text{total}}(\Phi_{t+1},\theta_t)]
&\le \mathbb E[\mathcal L_{\text{total}}(\Phi_t,\theta_t)] \\
&\quad - \eta\!\left(1-\tfrac{L_\Phi\eta}{2}\right)\mathbb E[\|\nabla_\Phi \mathcal L_{\text{total}}(\Phi_t,\theta_t)\|^2] \\
&\quad + \tfrac{L_\Phi\eta^2}{2}\sigma^2.
\end{aligned}
\]

Now account for the drift in $\theta_t$. Since $\mathcal L_{\text{total}}(\Phi,\theta)$ is $L_\theta$-Lipschitz in $\theta$, we add
\[
+ L_\theta\,\mathbb E|\theta_{t+1}-\theta_t|.
\]

Under the two-time-scale condition $\sum_t\mathbb E|\theta_{t+1}-\theta_t|<\infty$, the perturbation is summable. Hence
\[
\sum_t \eta\,\mathbb E[\|\nabla_\Phi \mathcal L_{\text{total}}(\Phi_t,\theta_t)\|^2] < \infty.
\]

Thus
\[
\liminf_{t\to\infty} \mathbb E[\|\nabla_\Phi \mathcal L_{\text{total}}(\Phi_t,\theta_t)\|]=0,
\]
and since the losses are lower-bounded and the weights bounded, $\mathbb E[\mathcal L_{\text{total}}(\Phi_t,\theta_t)]$ converges.
\end{proof}

\begin{table*}[t]
\centering
\captionsetup{skip=3pt}
\caption{Unsupervised node classification F1-score on clean datasets and under 30\% noise. 
The highest results are highlighted in \textbf{bold} and the second-best results are \underline{underlined}. 
ACC results are reported in Table~\ref{tab:unsupervised_acc_combined}.}
\label{tab:unsupervised_f1_combined}
\resizebox{1.0\linewidth}{!}{%
  \setlength{\tabcolsep}{4pt}
  \footnotesize
  \begin{tabular}{@{}l*{12}{c}@{}}
  \toprule
  \multirow{2}{*}{Methods} 
  & \multicolumn{2}{c}{Pubmed} & \multicolumn{2}{c}{Cora} & \multicolumn{2}{c}{Citeseer} & \multicolumn{2}{c}{WikiCS} & \multicolumn{2}{c}{Reddit} & \multicolumn{2}{c}{Instagram} \\
  \cmidrule(lr){2-3} \cmidrule(lr){4-5} \cmidrule(lr){6-7} \cmidrule(lr){8-9} \cmidrule(lr){10-11} \cmidrule(lr){12-13}
   & 0\% & 30\% & 0\% & 30\% & 0\% & 30\% & 0\% & 30\% & 0\% & 30\% & 0\% & 30\% \\
  \midrule
  GCN         & ${63.48}_{\pm 2.48}$ & -- & ${65.73}_{\pm 3.74}$ & -- & ${53.31}_{\pm 5.13}$ & -- & ${35.34}_{\pm 1.88}$ & -- & ${48.73}_{\pm 4.71}$ & -- & ${46.84}_{\pm 0.80}$ & -- \\
  GAT         & ${64.10}_{\pm 0.57}$ & -- & ${64.80}_{\pm 5.40}$ & -- & ${48.94}_{\pm 2.40}$ & -- & ${40.63}_{\pm 2.39}$ & -- & ${51.29}_{\pm 2.43}$ & -- & ${47.99}_{\pm 1.07}$ & -- \\
  DGI         & ${66.39}_{\pm 1.59}$ & -- & ${63.76}_{\pm 3.14}$ & -- & ${32.54}_{\pm 5.52}$ & -- & ${31.95}_{\pm 3.35}$ & -- & $\mathbf{62.99}_{\pm 0.55}$ & -- & ${49.32}_{\pm 0.66}$ & -- \\
  GRACE       & ${64.22}_{\pm 2.34}$ & -- & ${67.50}_{\pm 2.52}$ & -- & ${57.55}_{\pm 2.47}$ & -- & ${37.34}_{\pm 1.84}$ & -- & ${48.37}_{\pm 4.10}$ & -- & ${44.06}_{\pm 0.91}$ & -- \\
  GraphCL     & ${63.01}_{\pm 2.23}$ & -- & ${61.74}_{\pm 5.76}$ & -- & ${49.27}_{\pm 5.37}$ & -- & ${38.30}_{\pm 4.52}$ & -- & ${49.26}_{\pm 4.04}$ & -- & ${46.36}_{\pm 1.24}$ & -- \\ \midrule
  BERT        & ${39.73}_{\pm 0.61}$ & -- & ${41.62}_{\pm 1.97}$ & -- & ${44.41}_{\pm 1.95}$ & -- & ${34.48}_{\pm 0.61}$ & -- & ${51.35}_{\pm 1.26}$ & -- & ${56.87}_{\pm 0.28}$ & -- \\
  SBERT       & ${53.18}_{\pm 1.41}$ & -- & ${58.23}_{\pm 2.89}$ & -- & ${64.10}_{\pm 1.20}$ & -- & ${57.93}_{\pm 0.24}$ & -- & ${52.76}_{\pm 0.70}$ & -- & ${32.06}_{\pm 1.21}$ & -- \\
  RoBERTa     & ${37.12}_{\pm 1.24}$ & -- & ${41.86}_{\pm 1.19}$ & -- & ${38.73}_{\pm 2.88}$ & -- & ${35.67}_{\pm 1.07}$ & -- & ${51.54}_{\pm 0.87}$ & -- & ${52.27}_{\pm 0.90}$ & -- \\
  DeBERTa     & ${45.79}_{\pm 1.28}$ & -- & ${40.51}_{\pm 1.43}$ & -- & ${34.99}_{\pm 2.47}$ & -- & ${34.88}_{\pm 1.13}$ & -- & ${50.82}_{\pm 0.38}$ & -- & ${50.80}_{\pm 2.02}$ & -- \\
  Qwen3-0.6B  & ${47.66}_{\pm 5.29}$ & -- & ${47.93}_{\pm 1.90}$ & -- & ${43.70}_{\pm 3.26}$ & -- & ${40.35}_{\pm 4.10}$ & -- & ${51.93}_{\pm 0.79}$ & -- & ${54.85}_{\pm 1.06}$ & -- \\ \midrule
  G2P2        & ${48.21}_{\pm 9.85}$ & ${41.83}_{\pm 8.49}$ & ${46.41}_{\pm 4.96}$ & ${33.00}_{\pm 7.62}$ & ${20.97}_{\pm 2.93}$ & ${17.63}_{\pm 1.78}$ & ${8.16}_{\pm 6.87}$ & ${10.61}_{\pm 4.22}$ & ${36.60}_{\pm 3.45}$ & ${33.35}_{\pm 0.30}$ & ${39.24}_{\pm 12.62}$ & ${40.38}_{\pm 8.36}$ \\
  G2P2*(2-shots) & ${60.42}_{\pm 5.03}$ & ${51.11}_{\pm 4.07}$ & ${49.92}_{\pm 3.43}$ & ${42.16}_{\pm 4.97}$ & ${23.63}_{\pm 3.62}$ & ${20.58}_{\pm 3.15}$ & ${57.44}_{\pm 3.84}$ & ${55.39}_{\pm 3.19}$ & ${49.79}_{\pm 3.44}$ & ${49.54}_{\pm 1.24}$ & ${53.79}_{\pm 3.91}$ & ${54.09}_{\pm 3.83}$ \\
  G2P2*(8-shots) & ${70.19}_{\pm 2.48}$ & ${60.37}_{\pm 2.95}$ & ${56.71}_{\pm 3.54}$ & ${51.21}_{\pm 2.07}$ & ${28.48}_{\pm 2.48}$ & ${24.40}_{\pm 1.71}$ & ${62.76}_{\pm 1.46}$ & ${59.30}_{\pm 1.92}$ & ${50.24}_{\pm 3.94}$ & ${51.34}_{\pm 2.57}$ & ${52.68}_{\pm 3.76}$ & ${53.40}_{\pm 3.81}$ \\
  ConGraT     & ${55.28}_{\pm 4.61}$ & ${51.91}_{\pm 5.08}$ & ${20.31}_{\pm 2.31}$ & ${18.39}_{\pm 2.64}$ & ${20.84}_{\pm 3.26}$ & ${17.65}_{\pm 1.99}$ & ${10.11}_{\pm 2.35}$ & ${10.47}_{\pm 2.38}$ & ${53.11}_{\pm 1.32}$ & ${52.27}_{\pm 1.29}$ & ${49.38}_{\pm 3.56}$ & ${50.36}_{\pm 1.83}$ \\ \midrule
    \rowcolor{mygray}
  \model      & $\mathbf{77.12}_{\pm 2.15}$ & $\mathbf{75.65}_{\pm 1.64}$ & $\mathbf{72.47}_{\pm 3.94}$ & $\mathbf{70.54}_{\pm 1.55}$ & $\mathbf{70.64}_{\pm 2.45}$ & $\mathbf{67.58}_{\pm 2.18}$ & $\mathbf{64.06}_{\pm 1.19}$ & $\mathbf{62.80}_{\pm 2.86}$ & $\underline{56.16}_{\pm 1.52}$ & $\mathbf{55.82}_{\pm 1.50}$ & $\mathbf{59.89}_{\pm 1.57}$ & $\mathbf{58.05}_{\pm 1.86}$ \\
  \model-CLIP & ${73.72}_{\pm 2.99}$ & ${70.29}_{\pm 2.40}$ & ${68.81}_{\pm 3.46}$ & ${65.46}_{\pm 3.88}$ & ${66.17}_{\pm 2.07}$ & ${60.64}_{\pm 5.26}$ & ${60.65}_{\pm 3.77}$ & ${58.82}_{\pm 3.21}$ & ${46.25}_{\pm 1.93}$ & ${45.09}_{\pm 1.23}$ & ${43.07}_{\pm 1.31}$ & ${43.39}_{\pm 2.39}$ \\
  w/o Assessment & ${73.26}_{\pm 3.34}$ & ${72.85}_{\pm 3.59}$ & ${69.14}_{\pm 2.45}$ & ${66.76}_{\pm 3.91}$ & ${67.53}_{\pm 2.31}$ & ${64.01}_{\pm 1.78}$ & ${62.81}_{\pm 1.65}$ & ${60.09}_{\pm 2.57}$ & ${55.06}_{\pm 3.24}$ & ${53.70}_{\pm 2.45}$ & ${58.36}_{\pm 1.41}$ & ${56.42}_{\pm 1.53}$ \\
  w/o Filter & $\underline{76.28}_{\pm 2.78}$ & $\underline{73.16}_{\pm 0.67}$ & $\underline{70.40}_{\pm 3.31}$ & $\underline{67.36}_{\pm 1.10}$ & $\underline{68.61}_{\pm 1.41}$ & $\underline{64.87}_{\pm 2.82}$ & $\underline{63.70}_{\pm 2.94}$ & $\underline{60.99}_{\pm 2.99}$ & ${55.28}_{\pm 1.97}$ & $\underline{53.76}_{\pm 2.50}$ & $\underline{58.52}_{\pm 1.96}$ & $\underline{57.44}_{\pm 2.81}$ \\ 
  \bottomrule
  \end{tabular}%
}
\end{table*}

\section{Experimental Setup Details}
\label{app:experimental_setup_details}

\subsection{Datasets}
\label{app:datasets_in_setup}
We evaluate on widely used Text-Attributed Graphs (TAGs) from \textbf{GLBench}~\cite{li2024glbench} and \textbf{GraphCLIP}~\cite{zhu2025graphclip}, covering citation networks, web graphs, and social networks. \textbf{Unsupervised setting:}  
Cora~\cite{sen2008collective} (2.7K papers, 7 classes), Citeseer~\cite{giles1998citeseer} (3.2K papers, 6 classes), Pubmed~\cite{yang2016revisiting} (19.7K biomedical papers, 3 classes), WikiCS~\cite{mernyei2020wiki} (11.7K CS articles, 10 classes), Reddit~\cite{GraphAdapter} (33K users, reply network, 2 classes), and Instagram~\cite{GraphAdapter} (11K users, following graph, 2 classes). \textbf{Transfer setting:}  
Pre-train on Pubmed and Reddit; evaluate on Cora, Citeseer, WikiCS, Instagram, Ele-Photo, Ele-Computers, and Books-History~\cite{yan2023comprehensive}. These datasets span citation, social, and e-commerce graphs with diverse label spaces. All follow standardized processing pipelines. To simulate noise, we inject $30\%$ inter-class text mismatches in training sets while keeping validation/test splits clean.

\subsection{Baselines}
\label{app:baseline_details_in_setup}
We compare with three families of methods: (1) \textbf{Graph SSL:} GCN~\cite{gcn}, GAT~\cite{gat}, DGI~\cite{dgi}, GRACE~\cite{grace}, GraphCL~\cite{gcl}, BGRL~\cite{bgrl}, GraphMAE~\cite{graphmae}. These models learn graph embeddings from structure via contrastive or generative self-supervision. (2) \textbf{Text encoders:} BERT~\cite{devlin2019bert}, SBERT~\cite{reimers2019sbert}, RoBERTa~\cite{liu2019roberta}, DeBERTa~\cite{he2020deberta}, Qwen3~\cite{yang2025qwen3technicalreport}. They encode textual attributes alone for zero-shot classification. (3) \textbf{Graph–text aligners:} G2P2~\cite{wen2023augmenting}, ConGraT~\cite{brannon2024congrat}, GraphCLIP~\cite{zhu2025graphclip}. These dual-encoder models align graph and text modalities using CLIP-style objectives. For transferability evaluation, ZeroG~\cite{li2024zerog} is included as a recent benchmark focusing on cross-dataset generalization.  
All baselines follow official implementations with reported hyperparameters; when unspecified, we adopt default values (e.g., LR $1\!\times\!10^{-3}$ for GNNs, $2\!\times\!10^{-5}$ for PLMs).

\subsection{Evaluation Protocols}
\label{app:evaluation_protocols_in_setup}
Node classification uses cosine similarity between node and class embeddings (multimodal/text models) or clustering + Hungarian matching (graph SSL). Transfer setting uses a fixed SBERT for class embeddings.  
Link prediction employs inner-product scoring with AUC as a metric.  
Retrieval (Node$\leftrightarrow$Text) is evaluated with MRR and Recall@K.  

\begin{table*}[htbp]
\centering
\captionsetup{skip=3pt}
\caption{Zero-shot node classification F1-score on clean datasets and under 30\% noise. 
The highest results are highlighted in \textbf{bold} and the second-best results are \underline{underlined}. 
ACC results are reported in Table~\ref{tab:transferability_acc}.}
\label{tab:app_zeroshot_f1_combined}
\resizebox{1.0\linewidth}{!}{%
  \setlength{\tabcolsep}{4pt}
  \footnotesize
  \begin{tabular}{@{}l*{14}{c}@{}}
  \toprule
  \multirow{2}{*}{Methods} 
  & \multicolumn{2}{c}{Cora} & \multicolumn{2}{c}{Citeseer} & \multicolumn{2}{c}{WikiCS} & \multicolumn{2}{c}{Instagram} & \multicolumn{2}{c}{Ele-Photo} & \multicolumn{2}{c}{Ele-Computers} & \multicolumn{2}{c}{Ele-History} \\
  \cmidrule(lr){2-3} \cmidrule(lr){4-5} \cmidrule(lr){6-7} \cmidrule(lr){8-9} \cmidrule(lr){10-11} \cmidrule(lr){12-13} \cmidrule(lr){14-15}
   & 0\% & 30\% & 0\% & 30\% & 0\% & 30\% & 0\% & 30\% & 0\% & 30\% & 0\% & 30\% & 0\% & 30\% \\
  \midrule
  DGI       & ${8.98}_{\pm 3.88}$  & --  & ${11.21}_{\pm 1.76}$ & -- & ${2.37}_{\pm 1.00}$  & --  & ${52.73}_{\pm 1.10}$ & -- & ${4.75}_{\pm 2.14}$  & --  & ${10.48}_{\pm 2.02}$ & -- & ${4.13}_{\pm 0.70}$ & -- \\
  GRACE     & ${8.63}_{\pm 2.80}$  & --  & ${9.97}_{\pm 3.96}$  & --  & ${11.99}_{\pm 1.07}$ & -- & ${52.78}_{\pm 1.93}$ & -- & ${4.93}_{\pm 2.89}$  & --  & ${4.76}_{\pm 2.32}$ & -- & ${4.06}_{\pm 1.60}$ & -- \\
  BGRL      & ${13.70}_{\pm 2.95}$ & -- & ${7.87}_{\pm 1.51}$  & --  & ${11.80}_{\pm 1.49}$ & -- & $\underline{58.76}_{\pm 2.75}$ & -- & ${3.07}_{\pm 2.90}$ & -- & ${6.37}_{\pm 1.76}$ & -- & ${5.01}_{\pm 2.79}$ & -- \\
  GraphMAE  & ${15.94}_{\pm 2.23}$ & -- & ${11.76}_{\pm 3.53}$ & -- & ${7.41}_{\pm 1.01}$  & --  & $\mathbf{60.87}_{\pm 1.49}$ & -- & ${1.81}_{\pm 0.68}$ & -- & ${7.46}_{\pm 1.11}$ & -- & ${6.87}_{\pm 1.49}$ & -- \\
  ZeroG     & ${42.15}_{\pm 1.97}$ & ${38.21}_{\pm 2.27}$ & ${28.94}_{\pm 2.37}$ & ${28.22}_{\pm 1.90}$ & ${17.22}_{\pm 2.13}$ & ${18.52}_{\pm 3.20}$ & ${40.02}_{\pm 1.54}$ & ${41.02}_{\pm 1.29}$ & ${30.28}_{\pm 1.19}$ & ${27.15}_{\pm 2.31}$ & ${27.19}_{\pm 1.87}$ & ${20.84}_{\pm 2.16}$ & ${16.26}_{\pm 1.32}$ & ${15.06}_{\pm 0.32}$ \\
  \midrule
  G2P2      & ${14.53}_{\pm 3.70}$ & ${9.43}_{\pm 5.00}$  & ${18.61}_{\pm 1.97}$ & ${15.00}_{\pm 1.58}$ & ${12.37}_{\pm 4.55}$ & ${7.58}_{\pm 4.48}$  & ${47.09}_{\pm 2.57}$ & ${48.52}_{\pm 0.49}$ & ${4.93}_{\pm 2.80}$  & ${4.49}_{\pm 1.64}$  & ${10.67}_{\pm 3.91}$ & ${6.38}_{\pm 2.57}$ & ${2.28}_{\pm 1.07}$ & ${2.85}_{\pm 1.67}$ \\
  GraphCLIP & ${52.28}_{\pm 1.23}$ & ${49.85}_{\pm 2.88}$ & ${44.10}_{\pm 2.51}$ & ${43.18}_{\pm 3.73}$ & ${19.48}_{\pm 1.55}$ & ${15.68}_{\pm 1.33}$ & ${41.95}_{\pm 0.73}$ & ${43.07}_{\pm 0.64}$ & ${12.97}_{\pm 1.56}$ & ${11.87}_{\pm 1.59}$ & ${23.03}_{\pm 0.63}$ & ${20.39}_{\pm 1.89}$ & ${21.52}_{\pm 0.70}$ & ${22.07}_{\pm 0.75}$ \\
  \midrule
    \rowcolor{mygray}
  \model 
            & $\mathbf{59.96}_{\pm 3.53}$ & $\mathbf{60.55}_{\pm 2.47}$ & $\mathbf{55.80}_{\pm 5.05}$ & $\mathbf{57.37}_{\pm 3.29}$ & $\mathbf{46.93}_{\pm 2.52}$ & $\mathbf{44.95}_{\pm 5.58}$ & ${51.22}_{\pm 1.33}$ & $\mathbf{50.39}_{\pm 1.39}$ & $\mathbf{44.85}_{\pm 3.12}$ & $\mathbf{40.63}_{\pm 3.94}$ & $\mathbf{43.68}_{\pm 2.78}$ & $\underline{38.83}_{\pm 1.44}$ & $\mathbf{24.23}_{\pm 0.68}$ & $\mathbf{24.65}_{\pm 1.43}$ \\
  \model-CLIP   
            & ${59.58}_{\pm 1.54}$ & ${57.58}_{\pm 1.63}$ & ${55.11}_{\pm 0.65}$ & ${54.69}_{\pm 0.90}$ & ${31.26}_{\pm 3.08}$ & ${33.74}_{\pm 3.41}$ & ${44.23}_{\pm 0.64}$ & ${46.50}_{\pm 0.59}$ & ${18.28}_{\pm 1.38}$ & ${16.38}_{\pm 1.86}$ & ${31.20}_{\pm 0.70}$ & ${32.23}_{\pm 1.37}$ & ${23.44}_{\pm 0.31}$ & ${23.29}_{\pm 0.37}$ \\
  w/o Assessment 
            & ${58.28}_{\pm 0.67}$ & ${58.65}_{\pm 1.11}$ & ${55.54}_{\pm 2.06}$ & $\underline{56.31}_{\pm 1.58}$ & ${40.58}_{\pm 3.79}$ & $\underline{44.69}_{\pm 3.63}$ & ${44.35}_{\pm 0.93}$ & $\underline{47.04}_{\pm 1.37}$ & ${35.72}_{\pm 2.45}$ & $\underline{34.70}_{\pm 2.60}$ & ${34.12}_{\pm 1.37}$ & $\mathbf{40.10}_{\pm 1.47}$ & ${23.87}_{\pm 0.25}$ & ${23.81}_{\pm 0.24}$ \\
    w/o Filter 
            & $\underline{59.70}_{\pm 3.26}$ & $\underline{59.53}_{\pm 2.17}$ & $\underline{55.68}_{\pm 4.24}$ & ${52.64}_{\pm 3.39}$ & $\underline{42.08}_{\pm 2.12}$ & ${43.75}_{\pm 3.94}$ & ${49.90}_{\pm 1.62}$ & ${45.53}_{\pm 1.90}$ & $\underline{43.84}_{\pm 3.95}$ & ${34.67}_{\pm 3.83}$ & $\underline{43.53}_{\pm 1.44}$ & ${37.51}_{\pm 1.92}$ & $\underline{23.92}_{\pm 0.92}$ & $\underline{23.83}_{\pm 0.59}$ \\
  \bottomrule
  \end{tabular}%
}
\end{table*}

\begin{table*}[t]
\centering
\captionsetup{skip=3pt}
\caption{Complete results of retrieval tasks on clean (0\% noise) and noisy (30\% noise) datasets.}
\label{tab:retrieval_full_combined}
\resizebox{\linewidth}{!}{
\begin{tabular}{@{}lll*{5}{c}*{5}{c}@{}} 
\toprule
\multirow{2}{*}{Scope} & \multirow{2}{*}{Task} & \multirow{2}{*}{Methods} 
& \multicolumn{5}{c}{Clean (0\% noise)} 
& \multicolumn{5}{c}{Noisy (30\% noise)} \\ 
\cmidrule(lr){4-8} \cmidrule(lr){9-13}
& & & MRR & R@1 & R@5 & R@10 & R@20 
    & MRR & R@1 & R@5 & R@10 & R@20 \\ 
\midrule
\multirow{4}{*}{Global} 
 & \multirow{2}{*}{T2N} & GraphCLIP & ${86.47}_{\pm 5.17}$ & ${80.42}_{\pm 6.48}$ & ${93.97}_{\pm 3.65}$ & ${96.45}_{\pm 2.73}$ & ${98.04}_{\pm 1.69}$ 
                                        & ${72.16}_{\pm 9.57}$ & ${63.48}_{\pm 9.42}$ & ${82.73}_{\pm 9.82}$ & ${88.17}_{\pm 8.02}$ & ${92.06}_{\pm 6.33}$ \\
 &                        & \model    & ${90.40}_{\pm 2.89}$ & ${85.59}_{\pm 4.15}$ & ${96.46}_{\pm 1.33}$ & ${98.08}_{\pm 0.83}$ & ${99.06}_{\pm 0.45}$ 
                                        & ${81.40}_{\pm 6.01}$ & ${74.39}_{\pm 7.61}$ & ${90.12}_{\pm 4.05}$ & ${93.65}_{\pm 2.73}$ & ${95.95}_{\pm 1.82}$ \\
 \cmidrule(l){2-13}
 & \multirow{2}{*}{N2T} & GraphCLIP & ${5.95}_{\pm 1.60}$ & ${3.64}_{\pm 1.05}$ & ${7.85}_{\pm 2.10}$ & ${10.49}_{\pm 2.68}$ & ${14.08}_{\pm 3.57}$ 
                                        & ${3.18}_{\pm 0.84}$ & ${1.87}_{\pm 0.64}$ & ${4.15}_{\pm 1.04}$ & ${5.80}_{\pm 1.18}$ & ${8.02}_{\pm 1.62}$ \\
 &                        & \model    & ${36.68}_{\pm 3.50}$ & ${28.02}_{\pm 2.95}$ & ${46.00}_{\pm 4.29}$ & ${54.11}_{\pm 4.74}$ & ${62.54}_{\pm 4.50}$ 
                                        & ${18.60}_{\pm 4.01}$ & ${12.41}_{\pm 2.92}$ & ${24.68}_{\pm 5.28}$ & ${31.47}_{\pm 6.34}$ & ${39.56}_{\pm 7.44}$ \\
\midrule
\multirow{4}{*}{Category} 
 & \multirow{2}{*}{T2N} & GraphCLIP & ${91.19}_{\pm 3.60}$ & ${86.42}_{\pm 4.90}$ & ${97.05}_{\pm 1.98}$ & ${98.45}_{\pm 1.18}$ & ${99.22}_{\pm 0.63}$ 
                                        & ${80.79}_{\pm 8.53}$ & ${72.89}_{\pm 10.36}$ & ${90.70}_{\pm 6.27}$ & ${94.45}_{\pm 4.29}$ & ${96.81}_{\pm 2.84}$ \\
 &                        & \model    & ${93.50}_{\pm 1.76}$ & ${89.77}_{\pm 2.65}$ & ${98.08}_{\pm 0.74}$ & ${99.06}_{\pm 0.40}$ & ${99.57}_{\pm 0.24}$ 
                                        & ${86.75}_{\pm 4.26}$ & ${80.84}_{\pm 5.84}$ & ${94.21}_{\pm 2.21}$ & ${96.63}_{\pm 1.41}$ & ${98.07}_{\pm 0.80}$ \\
 \cmidrule(l){2-13}
 & \multirow{2}{*}{N2T} & GraphCLIP & ${17.52}_{\pm 2.41}$ & ${10.74}_{\pm 1.57}$ & ${23.13}_{\pm 3.48}$ & ${31.74}_{\pm 4.29}$ & ${42.55}_{\pm 4.95}$ 
                                        & ${11.82}_{\pm 1.82}$ & ${6.52}_{\pm 1.33}$ & ${15.83}_{\pm 2.04}$ & ${23.17}_{\pm 2.71}$ & ${32.86}_{\pm 3.89}$ \\
 &                        & \model    & ${51.50}_{\pm 4.26}$ & ${40.69}_{\pm 3.94}$ & ${63.82}_{\pm 5.05}$ & ${72.91}_{\pm 4.15}$ & ${81.23}_{\pm 3.81}$ 
                                        & ${32.56}_{\pm 5.48}$ & ${22.78}_{\pm 4.66}$ & ${42.97}_{\pm 6.78}$ & ${52.79}_{\pm 7.37}$ & ${63.22}_{\pm 7.33}$ \\
\bottomrule
\end{tabular}
}
\end{table*}

\subsection{Training Hyperparameters}
\label{app:training_hyper}

All experiments followed a consistent training setup unless otherwise specified. 
We trained for up to 20 epochs with a batch size of 16, learning rate $1.32\times10^{-4}$, weight decay $0.013$, and contrastive temperature $\tau=0.379$, using AdamW with a warmup ratio of 0.19 and early stopping (patience 5). 
Results were averaged over 10 seeds. 
For \model-specific parameters, we set the base weights to $\beta_0=0.5$, $\mu_0=2.0$, and $\gamma_0=0.5$, with EMA momentum $m=0.9$ and sensitivity $\alpha=0.7$. 
In fixed-$\theta$ experiments, $\theta$ was clamped to $\{0.9,1.0,1.1\}$, which also determined the corresponding sample filtering intensity. 
Unless otherwise noted, $\theta$ was initialized at 1.0 and allowed to adapt dynamically. 
Evaluation was conducted under both clean and noisy (30\% mismatch) settings, with performance reported on clean test sets. Experiments ran on a single NVIDIA A800.

\section{Additional Results for Experiments}
\label{app:Experiments_full_results}

\subsection{Unsupervised Setting}
\label{app:Unsupervised Setting}

Table~\ref{tab:unsupervised_f1_combined} reports F1-scores under both clean and noisy conditions, complementing the ACC results in the main text. Overall, \model consistently outperforms both unimodal and multimodal baselines across datasets. Compared with unimodal graph encoders, which may achieve competitive ACC by leveraging clean graph structure, \model achieves clearly higher F1-scores, reflecting a better balance between precision and recall even under noisy alignments. This highlights the benefit of combining many-to-many semantic alignment with dynamic filtering of noisy pairs.

Table~\ref{tab:unsupervised_f1_combined} also includes ablation results. \model substantially outperforms \model-CLIP, demonstrating the value of our dynamic framework beyond a standard CLIP-style loss. Moreover, both dynamic assessment and filtering are indispensable: removing either consistently reduces performance across clean and noisy datasets.

\subsection{Transfer Setting}
\label{app:gfm_full_results_combined}
\subsubsection{Zero-Shot Node Classification Results}
\label{app:app_zero_shot}

Tables~\ref{tab:transferability_acc} and \ref{tab:app_zeroshot_f1_combined} show that \textbf{\model} consistently improves upon GraphCLIP and other baselines across diverse datasets. Even when GraphCLIP achieves slightly higher ACC (e.g., on Instagram), \textbf{\model} yields notably better F1-scores, indicating more balanced predictions. This highlights the effectiveness of its dynamic loss in capturing richer semantic relationships beyond one-to-one alignment. Under 30\% noise, the advantages of \textbf{\model} are amplified: it maintains high ACC and F1 while other multimodal methods degrade substantially. Notably, \textbf{\model}'s gap over GraphCLIP widens on challenging datasets like WikiCS and Ele-Computers, confirming its robustness to noisy supervision.  

Ablation studies further validate that both adaptive filtering and adaptive loss objectives are essential. While \textbf{\model-CLIP} underperforms across the board, \textbf{w/o Filter} can even trail \textbf{w/o Assessment} under noise, since strict loss adjustments applied without dynamic filtering risk reinforcing noisy pairs. This underscores the importance of their synergy in enabling robust transfer.  

\subsubsection{Retrieval Results.}
\label{app:app_retrieval}
As shown in Table~\ref{tab:retrieval_full_combined}, integrating \textbf{\model}'s dynamic loss substantially enhances retrieval, with particularly large gains for the more challenging N2T task. 
Compared with GraphCLIP, \textbf{\model} achieves markedly higher MRR and Recall@K across both global and category-specific retrieval. 
These improvements hold under both clean and noisy pre-training, with the gap becoming even more pronounced in noisy settings, demonstrating that the dynamic loss not only strengthens fine-grained graph–text alignment but also improves robustness against noisy supervision.



\begin{figure*}[t]
    \centering
    \captionsetup{skip=3pt}
    \includegraphics[width=0.24\linewidth]{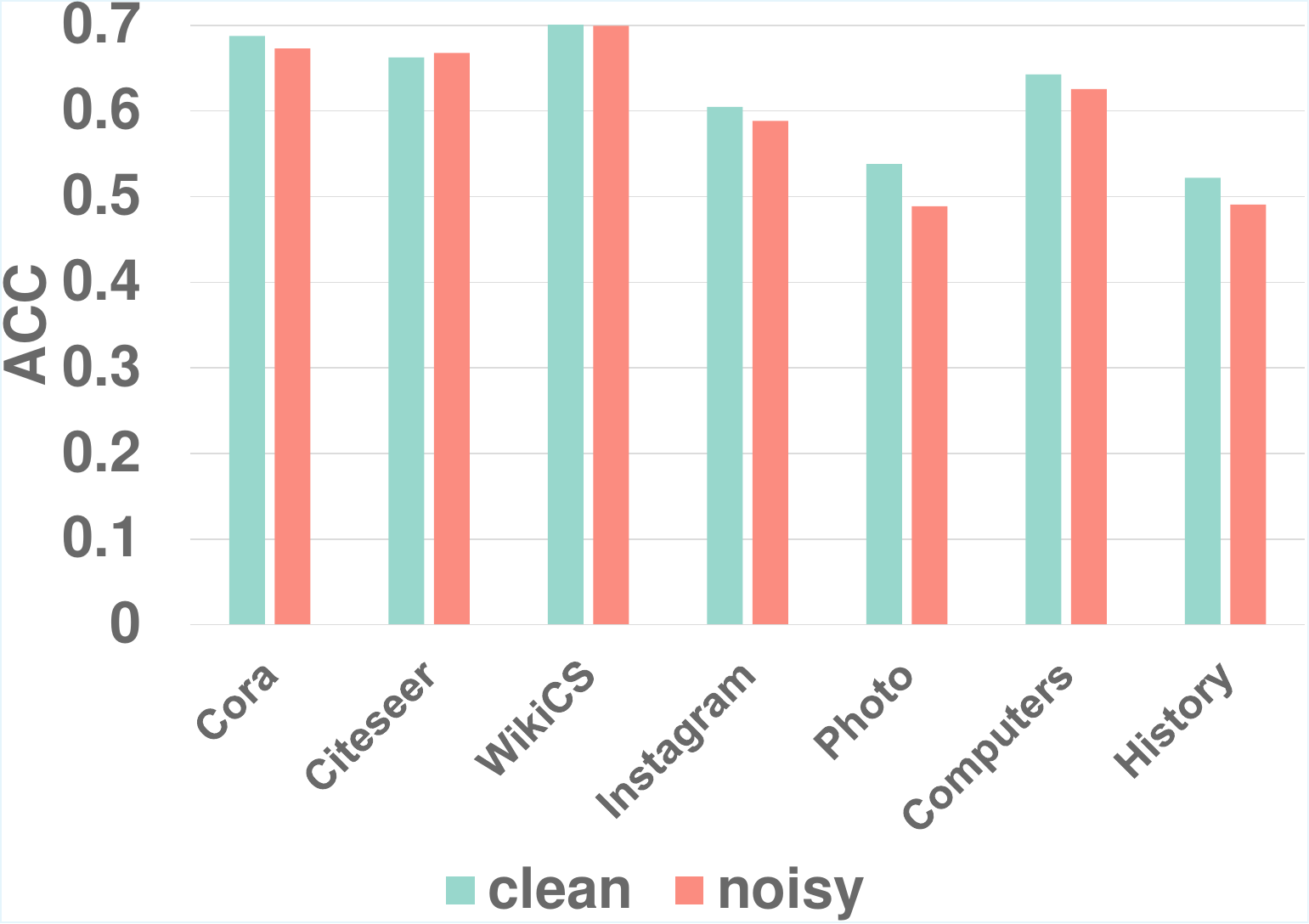}
    \includegraphics[width=0.24\linewidth]{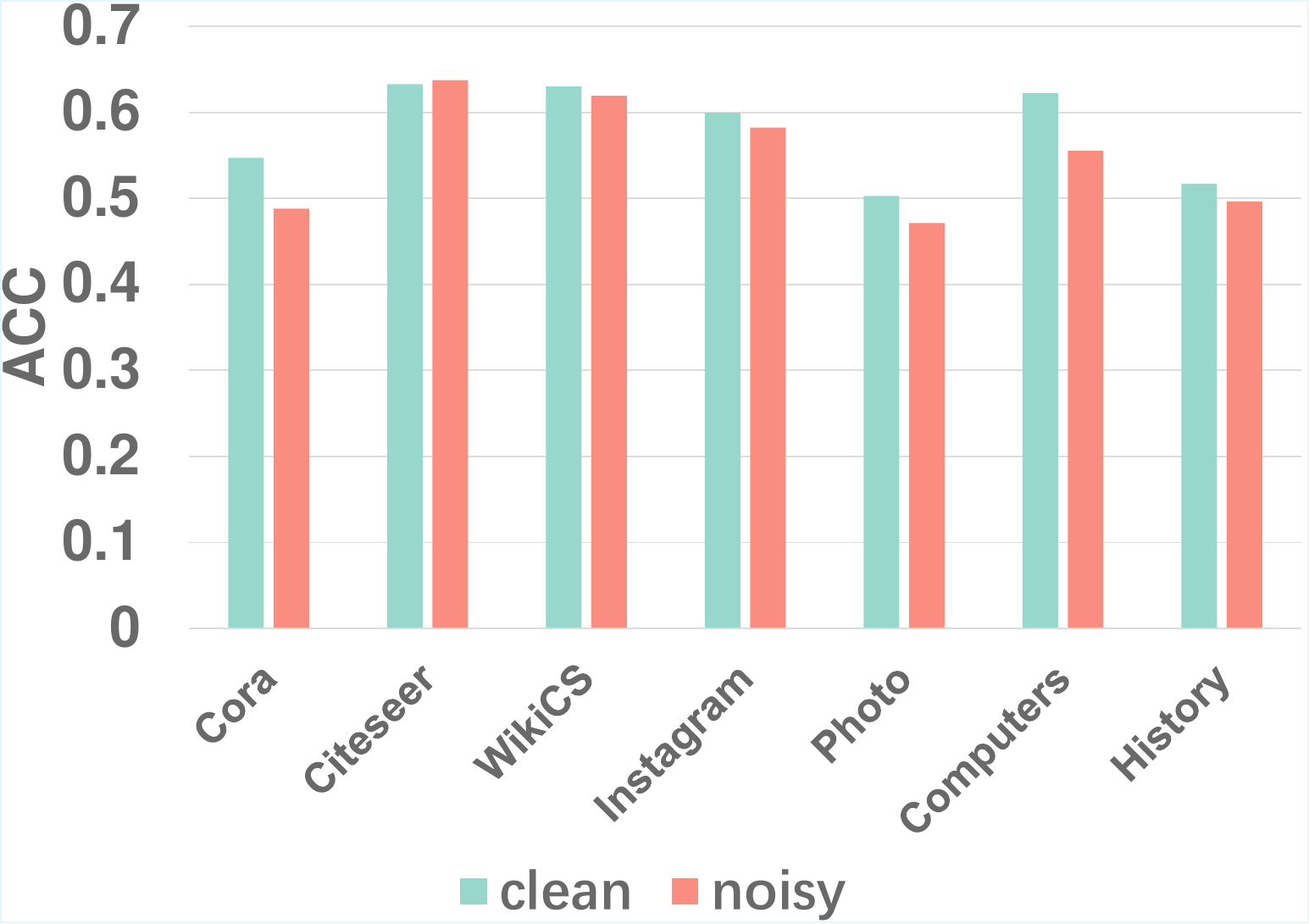}
    \includegraphics[width=0.24\linewidth]{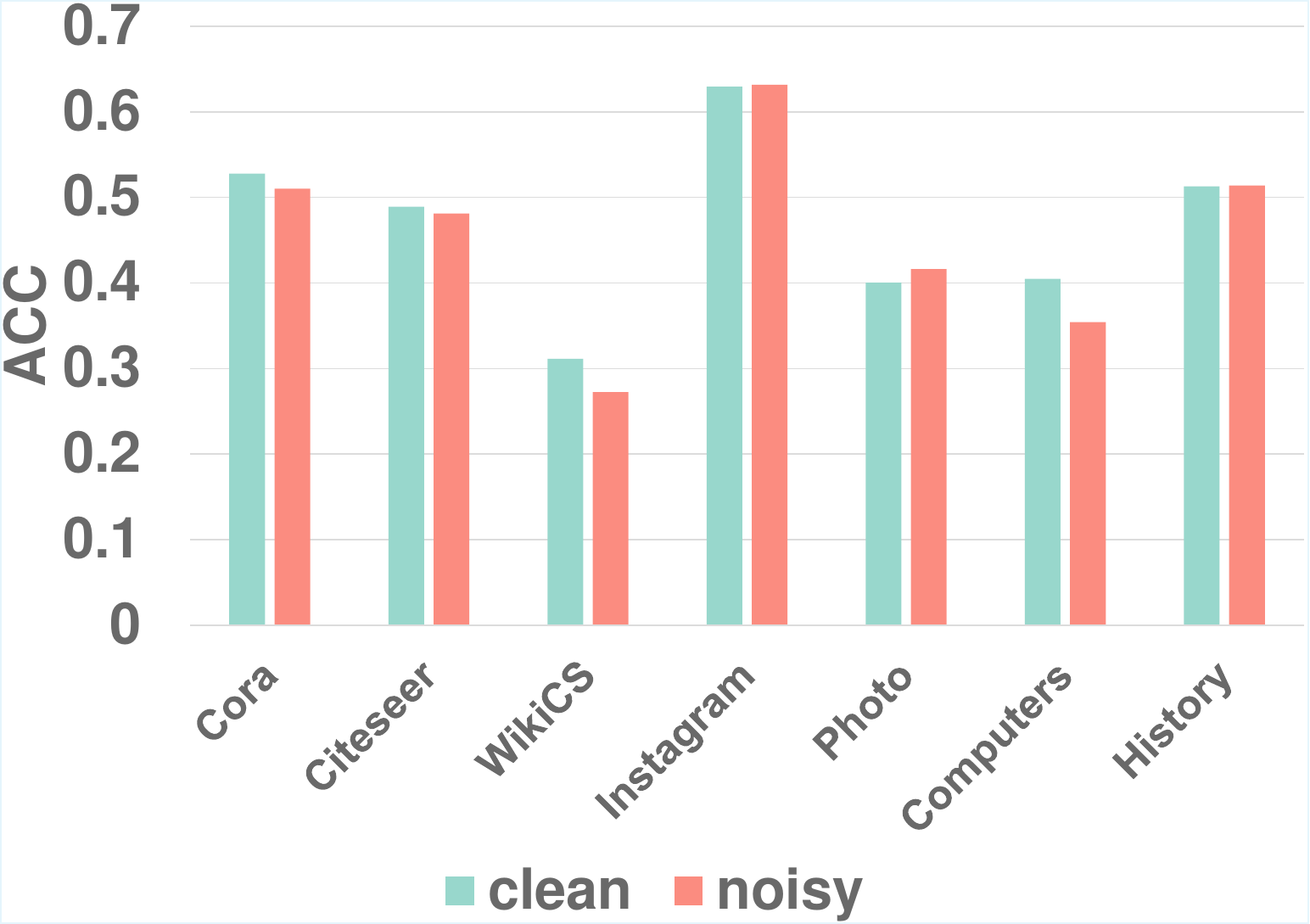}
    \includegraphics[width=0.24\linewidth]{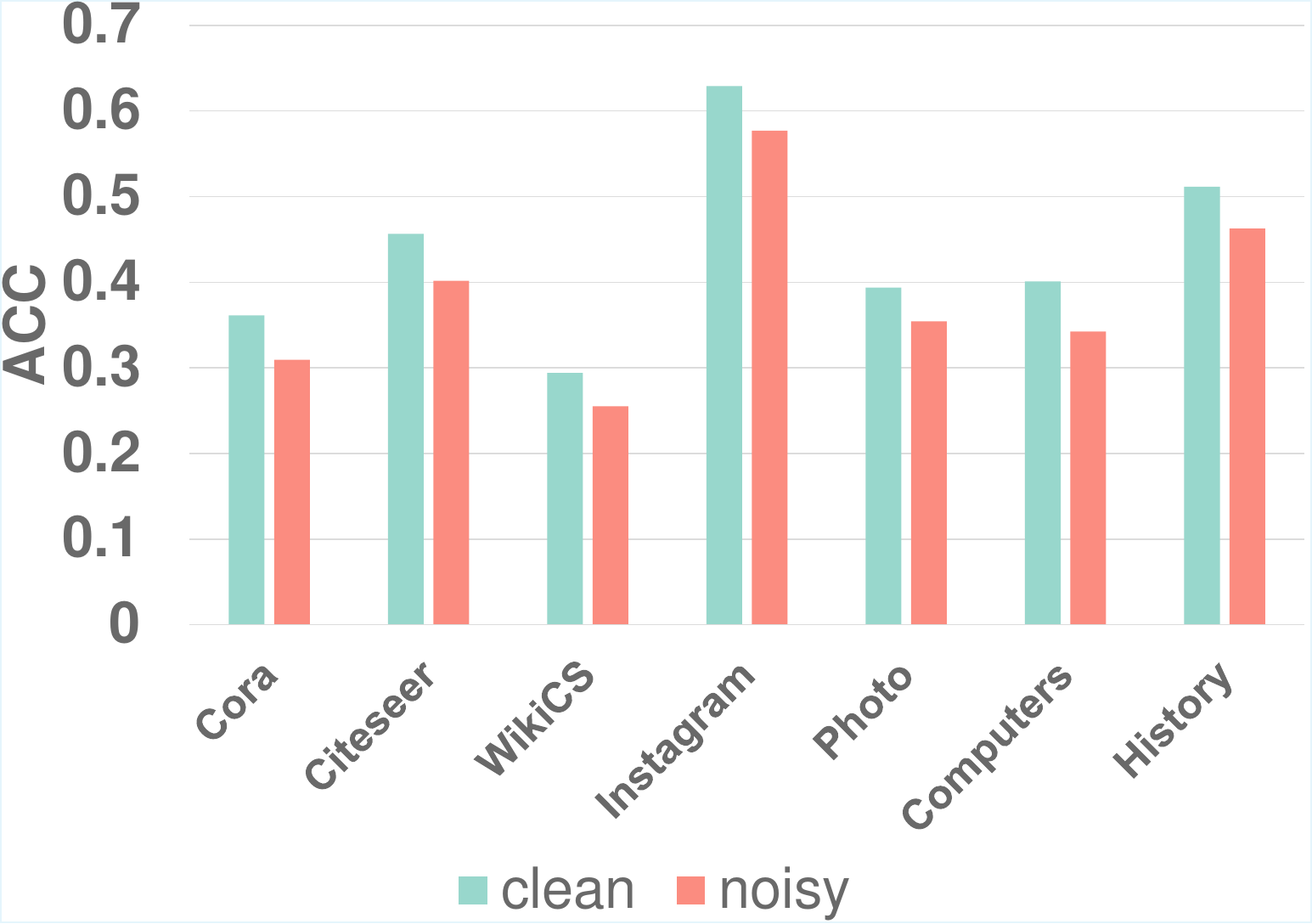}
    \caption{Zero-shot node classification performance on target datasets when pre-training with different source datasets: (a) all available sources, (b) Pubmed+Products+Reddit, (c) Pubmed+Reddit, and (d) Reddit only. Each figure compares models trained on clean versus noisy (30\% mismatch) data.}
    \label{fig:app_source_selection}
\end{figure*}

\subsubsection{Impact of Source Dataset Selection}
\label{app:app_source_dataset_selection}

Our supplementary analysis shows that the choice of source datasets in pre-training substantially affects zero-shot transfer performance and robustness to noise (Figures~\ref{fig:app_source_selection}). 
\noindent \textbf{Scale and Diversity.} 
Pre-training on larger and more diverse datasets generally improves robustness, as richer priors make the model less sensitive to noisy graph-text alignments and better at generalizing across domains. 
\noindent \textbf{Noise as Augmentation.} 
In some cases, limited amounts of mismatch noise even acted like data augmentation, slightly improving transfer when the source corpus was already large and diverse or when there was a large domain gap to the target task. However, this effect is inconsistent and cannot be relied upon as a robust learning strategy. 
\noindent \textbf{Limitations of Small-Scale Sources.} 
When source datasets are small or domain-limited, the GFM becomes highly vulnerable to noisy alignments, leading to significant performance drops. This underscores the need for a principled robust loss design. Our proposed dynamic quality-aware objective addresses this by explicitly identifying and mitigating noisy signals, ensuring stable representation learning even under constrained pre-training conditions.

\end{document}